\pdfoutput=1

\documentclass[11pt]{article}

\usepackage{EMNLP2022}

\usepackage{times}
\usepackage{latexsym}
\usepackage{graphicx}
\usepackage{color,soul}
\usepackage{hyperref}
\usepackage{amsmath}
\usepackage{subfigure}
\usepackage{hyperref}
\usepackage[shortlabels]{enumitem}
\usepackage[font=small,skip=1pt]{caption}

\DeclareMathAlphabet      {\mathbfit}{OML}{cmm}{b}{it}

\usepackage[T1]{fontenc}

\usepackage[utf8]{inputenc}

\usepackage{microtype}

\usepackage{amsmath}
\usepackage{amssymb}

\usepackage{caption}

\usepackage{booktabs}
\usepackage{multirow}

\title{Diving Deep into Modes of Fact Hallucinations in Dialogue Systems}

\author{Souvik Das \space Sougata Saha \space Rohini K. Srihari \\
        \texttt{\{souvikda, sougatas, rohini\}@buffalo.edu} \\
        Department of Computer Science and Engineering, University at Buffalo, NY. \\}

\begin{document}
\maketitle
\begin{abstract}
Knowledge Graph(KG) grounded conversations often use large pre-trained models and usually suffer from fact hallucination. Frequently entities with no references in knowledge sources and conversation history are introduced into responses, thus hindering the flow of the conversation—existing work attempt to overcome this issue by tweaking the training procedure or using a multi-step refining method. However, minimal effort is put into constructing an entity-level hallucination detection system, which would provide fine-grained signals that control fallacious content while generating responses. As a first step to address this issue, we dive deep to identify various modes of hallucination in KG-grounded chatbots through human feedback analysis. Secondly, we propose a series of perturbation strategies to create a synthetic dataset named \textsc{FADE} (\textbf{FA}ctual \textbf{D}ialogue Hallucination \textbf{DE}tection Dataset)\footnote{\url{https://github.com/souvikdgp16/FADE}}.  Finally, we conduct comprehensive data analyses and create multiple baseline models for hallucination detection to compare against human-verified data and already established benchmarks.
\end{abstract}

\section{Introduction}
Knowledge-grounded conversational models often use large pre-trained models \cite{gpt2, gpt3}. These models are notorious for producing responses that do not comply with the provided knowledge; this phenomenon is known as \textit{hallucination} \cite{dziri2, hall4}. Faithfulness to the supplementary knowledge is one of the prime designing factors in these knowledge-grounded chatbots. If a response is unfaithful to some given knowledge, it becomes uninformative and risks jeopardizing the flow of the conversation.  Despite retaining strong linguistics abilities, these large language models(LM) inadequately comprehend and present facts during conversations. LMs are trained to emulate distributional properties of data that intensify its hallucinatory attributes during test time.

\begin{figure}[h]
\centering
\includegraphics[width=0.4\textwidth]{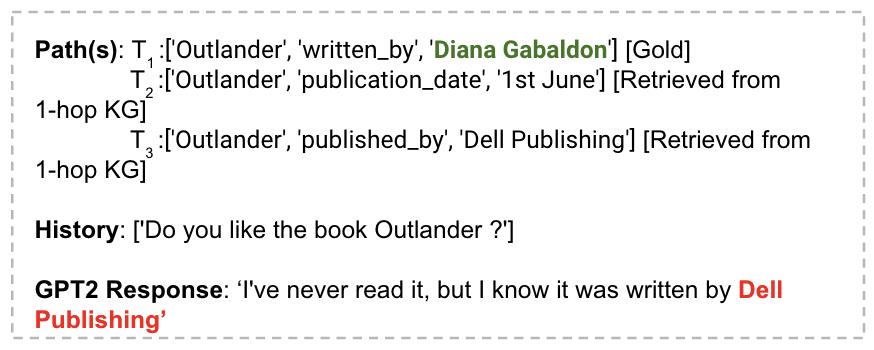}
\caption{Hallucination manifested by generated responses using GPT2\cite{gpt2} trained on KG triples can be more nuanced.}
\label{fig:example_anno}
\end{figure}
On the one hand, many prior works \cite{wiseman-etal-2017-challenges, parikh-etal-2020-totto, tuan-etal-2019-dykgchat} have suggested training these models on external data to ensure faithfulness may lead to a source-reference divergence problem, where the reference contains additional factual information. To address this problem holistically, \citeauthor{dziri-etal-2021-neural} has proposed a two-step generate-then-refine approach by augmenting conventional dialogue generation with a different refinement stage enabling the dialogue system to correct potential hallucinations by querying the KG. Also, this work employs a token-level hallucination classifier trained on a synthetic dataset constructed using two perturbation strategies \footnote{(1) Extrinsic perturbation: \citeauthor{dziri-etal-2021-neural} have swapped an entity with a different entity of the same type and not present in $1$-hop subgraph. (2) Intrinsic perturbation: they have swapped an entity with its object or vice versa, taken from the golden $1$-hop subgraph.}. Though this method has clear benefits, the hallucination perturbation strategies proposed in this work might fail to capture some of the subtle attributions of a factual generative model. As illustrated in Figure \ref{fig:example_anno}, neural models can inject hallucinated entities into responses that are present in the $k$-hop KG and are deceptively similar to what is expected. Also, if we cannot detect these elusive hallucinations beforehand, it will cause a cascading effect and amplify hallucinations in subsequent turns \cite{see-manning-2021-understanding}.   

On the other hand, relying on human annotations is challenging due to error-prone collection protocols and human ignorance to complete the tasks with care \cite{smith-etal-2022-human}. Prior research \cite{audit} shows that knowledge-grounded conversational benchmarks contain hallucinations promoted by a design framework that encourages informativeness over faithfulness. As studied by \citeauthor{dziri2}, when the annotators are asked to identify hallucination in a response, there is a high chance of error due to lack of incentive, personal bias, or poor attention to the provided knowledge. 

\citeauthor{see-manning-2021-understanding} have studied different shortcomings in a real-time neural model. In this work, based on some of the findings of \citeauthor{see-manning-2021-understanding}, like repetitive and unclear utterances promoting hallucination, we extend the already defined modes of hallucinations \cite{maynez-etal-2020-faithfulness, dziri-etal-2021-neural}.  Our contributions to this work are threefold:
\begin{itemize}[noitemsep,nolistsep]
  \item We extend fact hallucination in KG-grounded dialogue systems into eight categories. To understand the degree to which our defined classes exist in real-life data, we conduct a systematic human evaluation of data generated by a state-of-the-art neural generator.
  \item Since human annotation is expensive and often inaccurate, we design a series of novel perturbation strategies to simulate the defined ways of fact hallucinations and build a set of synthetic datasets collectively named as \textsc{FADE} (\textbf{FA}ctual \textbf{D}ialogue Hallucination \textbf{DE}tection Dataset).
  \item We create multiple pre-trained model-based baselines and compare the performances on several constituent and mixed datasets. To assess our dataset's generalization capability, we perform zero-shot inference on BEGIN \cite{begin}, and FaithDial \cite{faithdial} datasets, which encompasses all categories of hallucinated responses.
\end{itemize}
  
\section{Different Modes of Hallucination in KG-grounded Dialogue Systems} 
\subsection{Background}
We focus on the task of detecting hallucinated spans in dialogues that are factually grounded on factoids derived from multi-relational graphs $\mathcal{G}=(\mathcal{V}, \mathcal{E}, \mathcal{R})$, termed as Knowledge-Graphs(KG). Each KG consists of an directed edge triples $t= \left< \texttt{[SBJ], [PRE], [OBJ]}\right>$, where $\texttt{[SBJ], [OBJ]} \in \mathcal{V}$ are nodes denoting subject and object entities and $\texttt{[PRE]} \in \mathcal{R}$ is a predicate which can be understood as a relation type. Primarily, a neural dialogue system is guilty of generating hallucinated text when a valid path in the $k$-hop sub-graph  $ \mathcal{G}^{k}_{c} \in \mathcal{G} $ of the original KG anchored around a context entity $c$ does not support it.

Our study extends the work of \cite{dziri-etal-2021-neural} where they specifically explore two broad circumstances -- extrinsic and intrinsic to the provided KG,  under which LMs are likely to exhibit unfaithful behavior. Though this categorization is beneficial for detecting hallucinations, these categories can be further subdivided into subcategories, which are described in \S \ref{modes_of_halluc}.

\subsection{Base Dataset}
We use OpenDialKG \cite{moon-etal-2019-opendialkg}, a crowded-sourced English dialogue dataset where two workers are paired to chat about a particular topic(mainly movie, music, sport, and book). We use this dataset for training a GPT2-based model for generating data for human feedback analysis and creating the perturbed datasets. More details about the dataset can be found in \S \ref{opendialkg}

\subsection{Definitions} \label{modes_of_halluc}
\begin{figure*}
  \centering
    \includegraphics[width=0.85\textwidth]{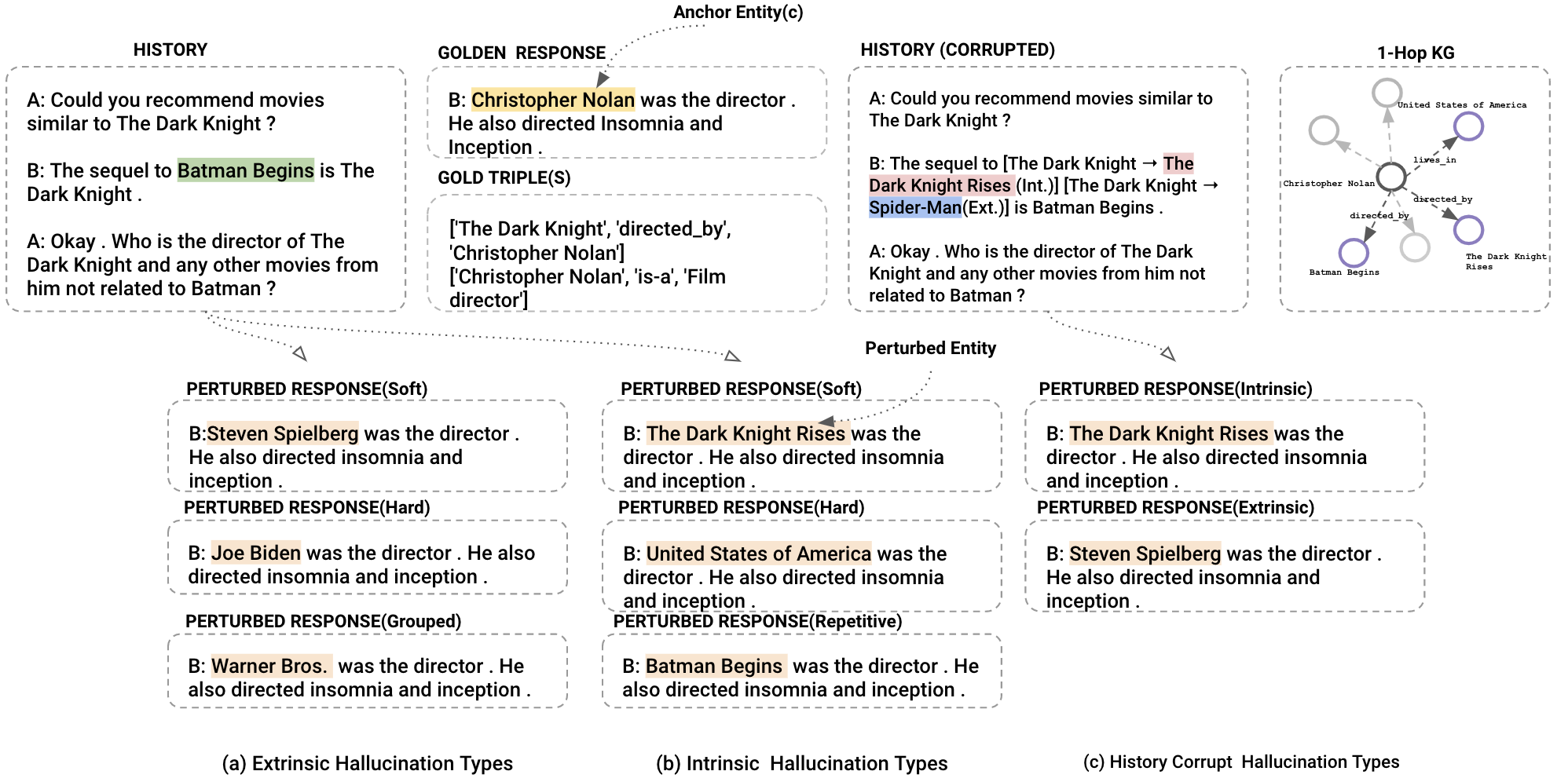}
    \caption{Illustration of our defined categories of fact hallucinations in KG-grounded dialogue systems}
    \label{fig:whole_schema}
\end{figure*}
We define below several categories of fact hallucination, comprehensive illustrations of each types are provided in Figure \ref{fig:whole_schema}. In addition we have included detailed descriptions of each definitions in \S \ref{sec:def_details}
\begin{enumerate}[(a), wide, labelwidth=!, labelindent=0pt, noitemsep, nolistsep]
  \item (\textbf{Extrinsic-Soft}). \textit{An extrinsic-soft hallucination corresponds to an utterance that brings a new span of text which is similar to the expected span but does not correspond to a valid triple in  $ \mathcal{G}^{k}_{c}$.}
  \item (\textbf{Extrinsic-Hard}). \textit{An extrinsic-hard hallucination corresponds to an utterance that brings a new span of text which is different from the expected span and does not correspond to a valid triple in  $ \mathcal{G}^{k}_{c}$.}
  \item (\textbf{Extrinsic-Grouped}). \textit{An extrinsic-grouped hallucination corresponds to an utterance that brings a new span of text which is different from the expected span but is of a specific predefined type and does not correspond to a valid triple in  $ \mathcal{G}^{k}_{c}$.}
  \item (\textbf{Intrinsic-Soft}). \textit{An intrinsic-soft hallucination corresponds to an utterance that misuses any triple in $ \mathcal{G}^{k}_{c}$ such that there is no direct path between the entities but they are similar to each other.}
  \item (\textbf{Intrinsic-Hard}). \textit{An intrinsic-hard hallucination corresponds to an utterance that misuses any triple in $ \mathcal{G}^{k}_{c}$ such that there is no direct path between the entities and they are not related in any form.}
  \item (\textbf{Intrinsic-Repetitive}). \textit{An intrinsic-repetitive hallucination corresponds to an utterance that either misuses \texttt{[SBJ]} or \texttt{[OBJ]} in $ \mathcal{G}^{k}_{c}$ such that there is no direct path between the entities but the entity has previously occurred in conversational history.}. 
  \item (\textbf{History Corrupted- Intrinsic/ Extrinsic}). \textit{A history corrupted(intrinsic/extrinsic) hallucination corresponds to an utterance that is subjected to intrinsic or extrinsic hallucination which is influenced by hallucinated entities in conversational history.}
\end{enumerate}
\subsection{Human Feedback Analysis} \label{sec:hu_feedback}
To study the extent to which the previously described modes of hallucination exist in a real-world system, we did human feedback analysis on responses generated using a GPT2-based generative model fine-tuned on OpenDialKG as described by \citeauthor{dziri-etal-2021-neural}. We sampled 200 responses each from four different decoding strategies, Greedy, Beam Search, and Nucleus Sampling, with a probability of 0.9 and 0.5. For each dialogue instance, we crowd-source human judgment by soliciting evaluations from 2 different annotators(with a high approval rating) from Amazon Mechanical Turk(AMT)(Details in \S \ref{sec:amt_ins}). One computer science graduate student additionally verified the Human Intelligence Task (HITS). For examples where hallucination was present, we asked the workers to identify the type of hallucination(examples of different types of hallucinations were shown in the instruction). The result of the human feedback is exhibited in Table \ref{tab:human_feedback}. We rejected $21\%$ of the HITS because of poor quality; we reported the average Krippendorf alpha coefficient to be 0.74 on the remaining annotations, indicating a moderate to a high agreement. Using Table \ref{tab:human_feedback} we made these observations:
\begin{itemize}[noitemsep,nolistsep]
  \item Extrinsic-soft hallucination is the dominant form of hallucination. Also, this bolsters our prior observation that LMs generate entities similar to the golden entity. 
  \item Comparatively less amount of hallucinations was seen in responses generated using beam search decoding scheme, though the percentage of extrinsic-hard hallucination was higher than greedy decoding. 
  \item Intrinsic-hard hallucination appears to be the least among all types. This suggests LM will always try to learn something from the given KG triples; generating something dissimilar will have a very low probability.
\end{itemize}

\begin{table}
\centering
\scalebox{0.5}{%
\begin{tabular}{@{}lrrrr@{}}
\toprule
\textbf{GPT2-KG} &
  \multicolumn{1}{l}{\textbf{Greedy}} &
  \multicolumn{1}{l}{\textbf{Beam Search}} &
  \multicolumn{1}{l}{\textbf{Nucleus 0.9}} &
  \multicolumn{1}{l}{\textbf{Nucleus 0.5}} \\ \midrule
Extrinsic-Soft              & 10.91 & 8.8   & 15.5  & 14.77 \\
Extrinsic-Hard              & 3.45  & 4.22  & 8.3   & 9.8   \\
Extrinsic-Grouped           & 1.12  & 1     & 0.44  & 1.6   \\
History Corrupted-Extrinsic & 3.3   & 3.1   & 2.33  & 1.1   \\ \midrule
Intrinsic-Soft              & 1.2   & 1.38  & 0.8   & 0.3   \\
Intrinsic-Hard              & 0.2   & 0.8   & 1.1   & 2     \\
Intrinsic-Repetitive        & 0.2   & 0.8   & 1.8   & 4     \\
History Corrupted-Intrinsic & 0.7   & 0.5   & 1.33  & 3.3   \\ \midrule
\textbf{Extrinsic Total}    & 18.78 & 17.12 & 26.57 & 27.27 \\
\textbf{Intrinsic Total}    & 2.3   & 3.48  & 5.03  & 9.6   \\ \midrule
\textbf{Total}              & 21.08 & 20.6  & 31.6  & 36.87 \\ \bottomrule
\end{tabular}}
\caption{Fine-grain human feedback analysis}
\label{tab:human_feedback}
\end{table}

\section{Dataset Creation}
\textsc{FADE} is a collection of datasets consisting of component datasets created using several perturbations and a set of mixed datasets constructed using the component datasets. 
\subsection{Perturbation Strategies}
\textbf{Extrinsic Hallucination} All the entities present in OpenDialKG undergo a indexing process. At first, using \textit{Spacy} we determine the named entity type \footnote{\url{https://spacy.io/api/entityrecognizer}} for each entity, and create BM25 indexes\footnote{\url{https://solr.apache.org/}} for each entity type. Each KG triple corresponding to an entity is represented in this format -- "\texttt{[SBJ]} \texttt{[PRE]} \texttt{[OBJ]}" and denoted as $t_i$. Now, for an entity($e_i$) we create a document $d_i = \mathrm{concat}(t_1, t_2, .. t_n)$, $n$ is the number of KG-triples for that entity. After this, we index $d_i$ and $e_i$ in the index corresponding to the entity type. During the perturbation process, we retrieve all the KG-triples for the entity we want to perturb and form 3 queries for each triple by permuting (\texttt{[SBJ]},\texttt{[PRE]},\texttt{[OBJ]}). Then based on the type of extrinsic hallucination, we query the indices to get the document scores in the following way: $\mathrm{scores} = \mathrm{average}(\{BM25(q_i, d_j)\}_{i\in{(s,r,o)}, j\in{(0,n)}})$, the selection criteria of the perturbed entities are provided in table \ref{tab:ext_sel}.

The groups for extrinsic-grouped hallucination are mentioned in Table \ref{tab:ent_group}. During the selection process, we iteratively check whether the perturbed entity exists in the conversation history, matches with the actual entity, and has appeared in the 1-hop sub-graph of the original entity. If an occurrence is found, we proceed to the following best entity.
\begin{table}[]
\centering
\scalebox{0.5}{%
\begin{tabular}{@{}lll@{}}
\toprule
\textbf{Hallucination Type} & \textbf{Index Type}                               & \textbf{Selection Criteria} \\ \midrule
Soft                        & Same as original entity                        & $e_i$ with max document score          \\
Hard                        & Same as original entity                        & $e_i$ with min document score          \\
Grouped                     & Same as one predefined type, selected randomly & same as soft          \\ \bottomrule
\end{tabular}}
\caption{Extrinsic hallucination perturbed entity selection criteria}
\label{tab:ext_sel}
\end{table}

\textbf{Intrinsic Hallucination} Here, we dynamically create a BM25 index and index all the KG triples in the 1-hop sub-graph of the original entity. Again, a KG triple is represented in the same fashion as in extrinsic hallucination -- "\texttt{[SBJ]} \texttt{[PRE]} \texttt{[OBJ]}". The goal here is to select entities that are similar or dissimilar to the original entities and present in the 1-hop graph. To achieve that, we follow a hybrid triple retrieval approach to score each triple associated with the original entity. First, we use the final hidden layer of a pre-trained GPT2 to obtain initial embeddings for each node in $ \mathcal{G}^{k}_{c}$ (for details, check \S \ref{kg_emb}). A query is formed by using Equation \ref{eq:query} each triple in $\mathcal{G}^{k}_{c}$ is scored using a similarity scoring system as described in Equation \ref{eq:sim}. 
{
\begin{equation} \label{eq:query}
   \mathbf{q} = \sum_{i\in \{s,r,o\}}^{}\frac{\varepsilon}{p(q_i)+\varepsilon}\ \mathbf{v_{q_i}}
\end{equation}
}
Here $\varepsilon$ is a free term parameter (\S \ref{int_perturb}), $p(q_i)$ is unigram probability of the query term and $\mathbf{v_{q_i}}$ is the embedding for each query term(here query terms are \texttt{[SBJ]}, \texttt{[PRE]} ,\texttt{[OBJ]} of the original entity).
{
\small
\begin{equation}\label{eq:ni}
   \mathbf{n_i} = \frac{\varepsilon }{p(s)+\varepsilon}\ \mathbf{v_{s}} + \frac{\varepsilon }{q(r)+\varepsilon}\ \mathbf{v_{r}}+ \frac{\varepsilon }{p(o)+\varepsilon}\ \mathbf{v_{o}}
\end{equation}
}
$\mathbf{n_i}$ in Equation \ref{eq:ni} represents a triple embedding in $\mathcal{G}^{k}_{c}$, when $q(r)$ represents the rarity of the relationship term in the subgraph, high occurrence is penalized, rest terms are analogous to Equation \ref{eq:query}.
{
\small
\begin{equation} \label{eq:sim}
   \mathrm{EntitySimilarity(Q,t)} = cos(\mathbf{q}, \mathbf{n_i})
\end{equation}
}

Now, we query the BM25 index that we have created before with a simple query using the original triple: "\texttt{[SBJ]} \texttt{[PRE]} \texttt{[OBJ]}" and get the score for each of the triple($t$). Finally, we get the final scores using Equation \ref{eq:fscore}.

{
\small
\begin{equation} \label{eq:fscore}
\begin{split}
   \mathrm{Score(Q,t)} = \beta \mathrm{EntitySimilarity(Q,t)} \\
   + (1 -\beta )\mathrm{BM25(Q,t)}
\end{split}
\end{equation}
}
Here $ 0< \beta < 1$.
\begin{table}[]
\centering
\scalebox{0.5}{%
\begin{tabular}{@{}ll@{}}
\toprule
\textbf{Hallucination Type} & \textbf{Selection Criteria}                                      \\ \midrule
Soft                        & \texttt{[SUB]} or \texttt{[OBJ]} with max triple score            \\
Hard                        & \texttt{[SUB]} or \texttt{[OBJ]} with min triple score           \\
Repetitive                  & same as soft, should be occurring in the conversation history \\ \bottomrule
\end{tabular}}
\caption{Intrinsic hallucination perturbed entity selection criteria}
\label{tab:int_sel}
\end{table}

We select the perturbed entities based on the scores and selection criteria as defined in Table \ref{tab:int_sel}. Like extrinsic hallucinations, we iteratively filter the best-scored entity until it does not match the original entity or appears in history. 

\textbf{History Corrupted Hallucination} Conversational history is corrupted using intrinsic or extrinsic corruption strategy. We select the last $k$ turns of the conversation and randomly perturb the entities. We also ensure that at least $50\%$ of the previous $k$ turns are corrupted.
\subsection{Dataset Analysis}
Below we provide data statistics and characterize the composition and properties of the datasets that are generated using our proposed perturbation strategies.
\subsubsection{Data Statistics}

\begin{table}[h]
\begin{minipage}[b]{60mm}
    \centering
    \scalebox{0.45}{%
    \begin{tabular}{@{}llll@{}}
    \toprule
    \textbf{Type} & \textbf{Perturbed} & \textbf{Non-perturbed} & \textbf{\begin{tabular}[c]{@{}l@{}}Turn with \\ perturbation\textgreater{}2\end{tabular}} \\ \midrule
    soft            & 12752 & 64634 & 558   \\
    hard            & 8540  & 68872 & 8254  \\
    grouped         & 22858 & 54542 & 11296 \\
    history-corrupt & 8534  & 68878 & 8247  \\ \bottomrule
    \end{tabular}}
    \caption{Extrinsic hallucination data statistics}
    \label{tab:ext_stats}
\end{minipage}

\begin{minipage}[b]{60mm}
    \centering
    \scalebox{0.45}{%
    \begin{tabular}{@{}llll@{}}
    \toprule
    \textbf{Type} & \textbf{Perturbed} & \textbf{Non-perturbed} & \textbf{\begin{tabular}[c]{@{}l@{}}Turn with \\ perturbation\textgreater{}2\end{tabular}} \\ \midrule
    soft            & 18560 & 58558 & 5 \\
    hard            & 18605 & 58534 & 6 \\
    repetitive      & 9712  & 67560 & 0 \\
    history-corrupt & 18597 & 58542 & 6 \\ \bottomrule
    \end{tabular}}
    \caption{Intrinsic hallucination data statistics}
    \label{tab:int_stats}
\end{minipage}

 \end{table}

Table \ref{tab:ext_stats} and \ref{tab:int_stats} shows the statistics of datasets created using different perturbation strategies. The base dataset contains 77,430 data points. However, the perturbed turns in each of these datasets are quite low in comparison. This low number is because not every entity in an utterance has a valid KG path. For extrinsic hallucination, $\sim$12,000 to $\sim$23,000 utterances were perturbed, and $\sim$550 to $\sim$11,300 utterances have multiple perturbations. The number of perturbed data points for intrinsic hallucination is less than extrinsic($\sim$9,000 to $\sim$18,000). The number of utterances with multiple perturbations is negligible due to the many checks the perturbed entities go through(for example, whether the KG path is present, has already occurred or not, etc.) To train and evaluate models, we vary the size of the train split in this range of $10\%$ to $30\%$\footnote{sequential split} with a step of $2.5\%$, keeping in mind to avoid overfitting. The remaining data is split into equal halves for validation and testing.
\subsubsection{Parsing Features}
\begin{figure}[h]
\centering
\includegraphics[width=0.4\textwidth]{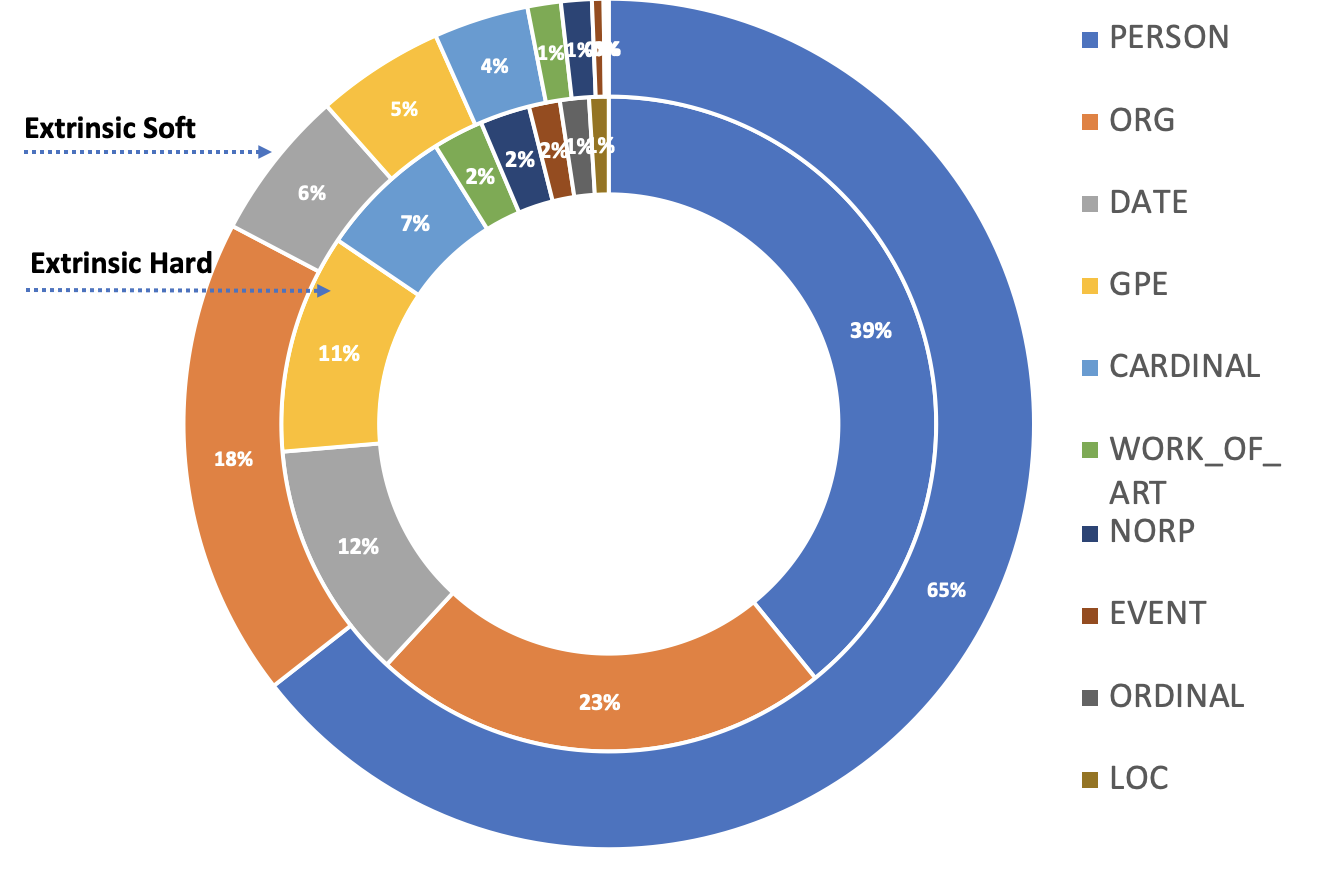}
\caption{NER distribution in Extrinsic-soft and hard hallucination}
\label{fig:ext_ner}
\end{figure}

\begin{figure}[h]
\centering
\includegraphics[width=0.4\textwidth]{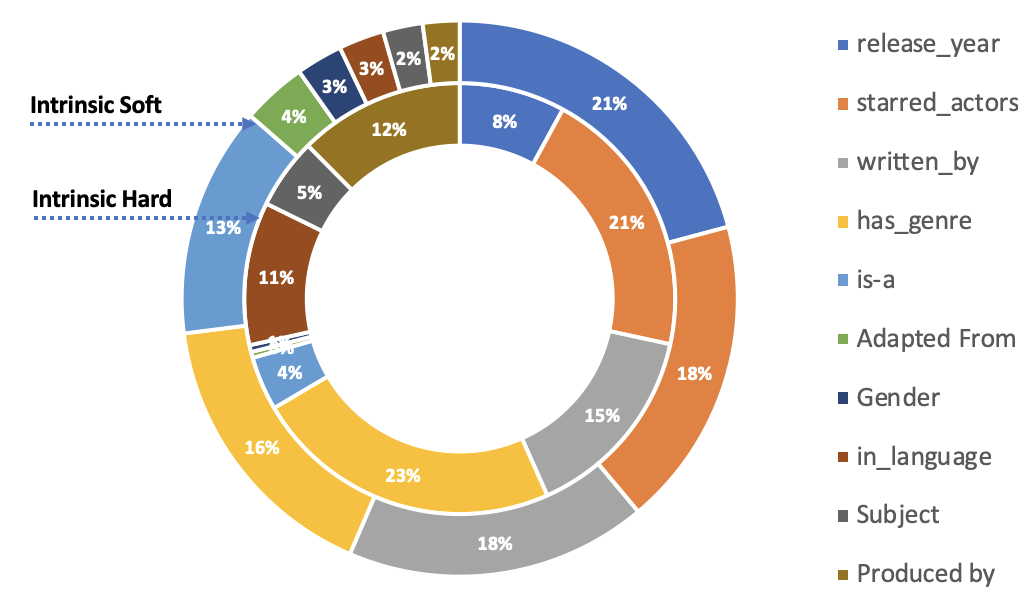}
\caption{Top 10 relation in perturbed KG triples in intrinsic-soft hallucination}
\label{fig:int_soft_rel}
\end{figure}

\begin{figure}[h]
\centering
\includegraphics[width=0.4\textwidth]{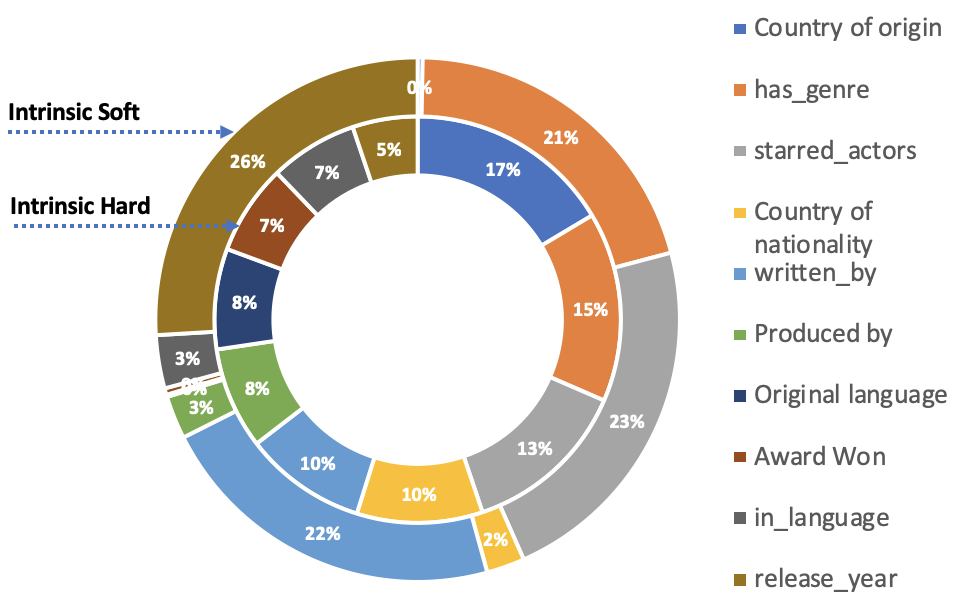}
\caption{Top 10 relation in perturbed KG triples in intrinsic-hard hallucination}
\label{fig:int_hard_rel}
\end{figure}

In Figure \ref{fig:ext_ner} we show the top 10 Named Entity Recognition(NER) tags as identified by the \textit{Spacy} library in extrinsic hallucinations. For extrinsic-soft hallucination, most NER tags are of type PERSON. This corresponds to the fact that the original entities in the base dataset are primarily related to movies, books, and music. In extrinsic-soft hallucination, the associated PERSON name is changed to a closely affiliated person, or a movie name is changed to its director's name. In contrast, the distribution of NER tags is uniform for extrinsic-hard hallucination.
Figure \ref{fig:int_soft_rel} and \ref{fig:int_hard_rel} shows the top-10 relations of the perturbed entity with the original entity in both intrinsic-soft and hard hallucinations and the corresponding value in their counterparts. In intrinsic-soft hallucination, more relevant relations are selected like "release year", "starred actors", "written by", etc. On the other hand, in intrinsic hard hallucination, more unusual relations like "Country of Origin", and "Country of Nationality" were among the top relations. 

\subsection{Mixing Datasets}
Since in actual data, all kinds of hallucinations are expected to occur. We mix the previously constructed datasets in specific proportions to create a more challenging dataset. Table \ref{tab:mixdata} shows the different mixing ratios for four types of datasets is as follows:  \textbf{Observed:} We try to mimic the observed data, which is shown in \S \ref{sec:hu_feedback}, we take an average of percentages in for all the decoding strategies. \textbf{Balanced:} Goal here is to create a balanced dataset between hallucinated and non-hallucinated turns, each type of hallucination is also balanced. \textbf{Extinsic+:} In this scenario, we increase the percentages of extrinsic-soft, hard, and grouped by a factor of 2, 1.5, and 1.5, respectively. \textbf{Intrinsic+:} here we increase the percentages of intrinsic-soft, hard and repetitive by a factor of 1.5. More details in \S \ref{mix_ratio}.

\subsection{Human Verification} \label{sec:hu_veri}
To verify whether our proposed perturbation strategies inject hallucinations in the original data, we randomly sample 150 examples from each of the mixed dataset's test splits. Subsequently, these samples were randomly ordered to form a consolidated sample of 600 data points annotated by at least three AMT workers, with the same setting as described in \S \ref{sec:hu_feedback}. Additionally, the graduate student verified where the hallucinations adhere to the perturbation norms. Krippendorff's alpha were 0.88 and 0.76 among workers, and workers with perturbed data(average), indicating a very high agreement. Since our perturbation strategies are purely deterministic, we kept a large-scale human verification of the automatically annotated data outside the scope of this work. We create a human-verified dataset of 500 samples, 300 taken from this set and 200 from the human feedback study \ref{sec:hu_feedback}.

\section{Task}
 To identify utterances that contain hallucinations and to locate the entities of concern. We create two tasks:
    \begin{enumerate}[noitemsep,nolistsep]
      \item \textbf{Utterance classification:} Given the dialog history $\mathcal{D}$, knowledge triples $\mathcal{K}_n$ and the current utterance $\overline{x}_{n+1}$ we classify $\overline{x}_{n+1}$ is hallucinated or not.
      \item \textbf{Token classification:} Given $\mathcal{D}$, $\mathcal{K}_n$ and $\overline{x}_{n+1}$, we need to perform sequence labelling on $\overline{x}_{n+1}$ and identify the hallucinated spans.
    \end{enumerate}

\section{Baseline Models}
As an initial effort toward tackling the suggested hallucination detection task, we create several baseline detection models based on pre-trained transformer models, including BERT, XLNet, and RoBERTa. These transformer-based models represent the state-of-the-art and can potentially better leverage context or embedded world knowledge to detect self-contradictory or anti-commonsense content.

For training the utterance classifier, given $\mathcal{D}$, $\mathcal{K}_n$ and $\overline{x}_{n+1}$, we fine tune a pre-trained model $\mathcal{M}$ to predict binary hallucinated label $\mathbf{y}$ for $\overline{x}_{n+1}$ . Here, $\mathcal{D}$ and $\mathcal{K}_n$ are considered as sequence A with token type ids as $0$ and $\overline{x}_{n+1}$ is considered as sequence B with token type ids as $1$. During inference, from the last hidden states $\mathbf{H} \in \mathbb{R}^{l \times h}$ ($h$, $l$ are hidden size and sequence length, respectively), then we obtain the representation $\mathbf{w} \in \mathbb{R}^{h}$ by max pooling(i.e., $\mathbf{w} =  max\_pool(\mathbf{H})$). We then pass $\mathbf{w}$ through a MLP layer with a $tanh$ activation to get the binary label $y \in \{0,1\}$. During training time, we fine-tune the model using cross entropy objective between the predicted labels and the actual labels.

Similarly, for training the sequence classifier, we fine-tune a pre-trained model $\mathcal{M}_s$. At first, we encode $\mathcal{D}$, $\mathcal{K}_n$ and $\overline{x}_{n+1}$ using $\mathcal{M}_s$ to get the last hidden states $\mathbf{H} \in \mathbb{R}^{l \times h}$, ($h$, $l$ are hidden size and sequence length, respectively). Instead of doing a binary classification of each token, we adopt a $BILOU$ encoding scheme. The hidden states are passed through an MLP layer with a $tanh$ activation to get the 5-way label $y \in \{B, I, L, O, U\}$. During training time, we fine-tune the model using a cross-entropy objective between the predicted and actual labels.

\section{Experimental Setup}

\begin{table*}[t]
\centering
\resizebox{12cm}{!}{%
\begin{tabular}{@{}l|l|lll|llllll@{}}
\toprule
\multicolumn{1}{c|}{\multirow{2}{*}{\textbf{Dataset}}} &
  \multicolumn{1}{c|}{\multirow{2}{*}{\textbf{Best Model}}} &
  \multicolumn{3}{c|}{\textbf{Token Level}} &
  \multicolumn{6}{c}{\textbf{Utterance Level}} \\ \cmidrule(l){3-11} 
\multicolumn{1}{c|}{} &
  \multicolumn{1}{c|}{} &
  \multicolumn{1}{c}{\textbf{F1}} &
  \multicolumn{1}{c}{\textbf{P}} &
  \multicolumn{1}{c|}{\textbf{R}} &
  \multicolumn{1}{c}{\textbf{F1}} &
  \multicolumn{1}{c}{\textbf{P}} &
  \multicolumn{1}{c}{\textbf{R}} &
  \multicolumn{1}{c}{\textbf{G-Mean}($\uparrow$)} &
  \multicolumn{1}{c}{\textbf{BSS}($\downarrow$)} &
  \multicolumn{1}{c}{\textbf{AUC}} \\ \midrule
extrinsic-grouped         & BERT(base-uncased) & 80.69 & 80.56 & 80.82 & 91.30 & 91.80 & 90.81 & 93.58 & 5.29 & 93.62 \\
extrinsic-hard            & XLNet(base-cased)  & 72.12 & 71.98 & 72.25 & 87.36 & 87.13 & 87.60 & 92.80 & 2.93 & 92.96 \\
extrinsic-history-corrupt & XLNet(base-cased)  & 72.38 & 72.35 & 72.40 & 88.10 & 87.86 & 88.34 & 93.24 & 2.75 & 93.38 \\
extrinsic-soft            & BERT(base-uncased) & 64.09 & 69.22 & 59.67 & 74.80 & 81.96 & 68.80 & 81.62 & 8.03 & 82.81 \\ \midrule
intrinsic-hard            & XLNet(base-cased)  & 84.44 & 85.08 & 83.81 & 90.88 & 92.88 & 88.97 & 93.24 & 4.48 & 93.34 \\
intrinsic-history-corrupt & XLNet(base-cased)  & 83.67 & 82.27 & 85.11 & 91.30 & 91.86 & 90.74 & 93.97 & 4.34 & 94.02 \\
intrinsic-repetitive      & RoBERTa(base)      & 82.70 & 82.76 & 82.64 & 88.01 & 89.51 & 86.55 & 92.31 & 3.15 & 92.50 \\
intrinsic-soft            & RoBERTa(base)      & 78.80 & 80.19 & 77.45 & 87.10 & 90.54 & 83.92 & 90.26 & 6.22 & 90.50 \\ \bottomrule
\end{tabular}%
}
\caption{Test benchmark (numbers in percentages (\%)) for component datasets, models trained on 25\% of the total dataset.}
\label{tab:ind_results}
\end{table*}

\begin{table*}[t]
\centering
\resizebox{10cm}{!}{%
\begin{tabular}{@{}l|l|lll|llllll@{}}
\toprule
\multicolumn{1}{c|}{\multirow{2}{*}{\textbf{Dataset}}} &
  \multicolumn{1}{c|}{\multirow{2}{*}{\textbf{Best Model}}} &
  \multicolumn{3}{c|}{\textbf{Token Level}} &
  \multicolumn{6}{c}{\textbf{Utterance Level}} \\ \cmidrule(l){3-11} 
\multicolumn{1}{c|}{} &
  \multicolumn{1}{c|}{} &
  \multicolumn{1}{c}{\textbf{F1}} &
  \multicolumn{1}{c}{\textbf{P}} &
  \multicolumn{1}{c|}{\textbf{R}} &
  \multicolumn{1}{c}{\textbf{F1}} &
  \multicolumn{1}{c}{\textbf{P}} &
  \multicolumn{1}{c}{\textbf{R}} &
  \multicolumn{1}{c}{\textbf{G-Mean}($\uparrow$)}&
  \multicolumn{1}{c}{\textbf{BSS}($\downarrow$)}&
  \multicolumn{1}{c}{\textbf{AUC}} \\ \midrule
\textbf{balanced} &
  RoBERTa-base &
  73.41 &
  68.75 &
  78.74 &
  88.24 &
  83.85 &
  93.12 &
  \textbf{86.21} &
  13.14 &
  \textbf{86.47} \\
\textbf{observed} &
  XLNet(base-cased) &
  63.44 &
  57.98 &
  70.03 &
  77.71 &
  71.05 &
  85.73 &
  85.40 &
  14.73 &
  85.40 \\
\textbf{intrinsic+} &
  RoBERTa-base &
  75.05 &
  71.11 &
  79.44 &
  90.16 &
  86.52 &
  94.12 &
  84.51 &
  12.78 &
  85.00 \\ 
  \textbf{extrinsic+} &
  XLNet(base-cased) &
  \textbf{75.59} &
  \textbf{70.79} &
  \textbf{81.10} &
  \textbf{90.75} &
  \textbf{86.77} &
  \textbf{95.11} &
  83.21 &
  \textbf{12.65} &
  83.95 \\ \bottomrule
\end{tabular}%
}
\caption{Test benchmark (numbers in percentages (\%)) for mixed datasets, models trained on 25\% of the total dataset.}
\label{tab:main_results}
\end{table*}

\textbf{Baseline configurations} we experiment with a variety of pre-trained models via Hugging Face Transformers, including \textbf{BERT}-base-uncased(110M), \textbf{RoBERTa}-base(125M) and \textbf{XLNet}-base-cased(110M). Though using large or medium versions of these models will produce better results, we refrain from using those models as scaling large models in production is costly. More details about training parameters can be found in \S \ref{impl_det}

We also experimented with model architecture as follows: \textit{(i)} Varied the length of the history \textit{(ii)} Experimented with max/ mean pooling. \textit{(iii)} Whether to concatenate the hidden states corresponding to  $\mathcal{K}_n$  with the hidden states corresponding to $\overline{x}_{n+1}$ before passing them through the MLP layer. \textit{(iv)} Using a CRF layer instead of MLP for predicting labels in the sequence tagger. The best configuration uses 4 turns of conversational history, max pooling, it does not concatenate hidden states of $\mathcal{K}_n$ with hidden states of $\overline{x}_{n+1}$ and uses a 2-layer MLP.

\textbf{Evaluation metrics} We evaluate the baselines with formal classification metrics, including precision, recall, and F1 for the hallucination sequence tagger. For the utterance-level hallucination classifier, we report accuracy, precision, recall, F1, and AUC (Area Under Curve) for ROC. We also use the G-Mean metric \cite{espindola2005extending}, which measures the geographic mean of sensitivity and specificity. We also employ the Brier Skill Score (BSS) metric \cite{center2005brier}, which computes the mean squared error between the reference distribution and the hypothesis probabilities.
\section{Results and Discussion}
\begin{figure*}
  \centering
    \includegraphics[width=0.9\textwidth]{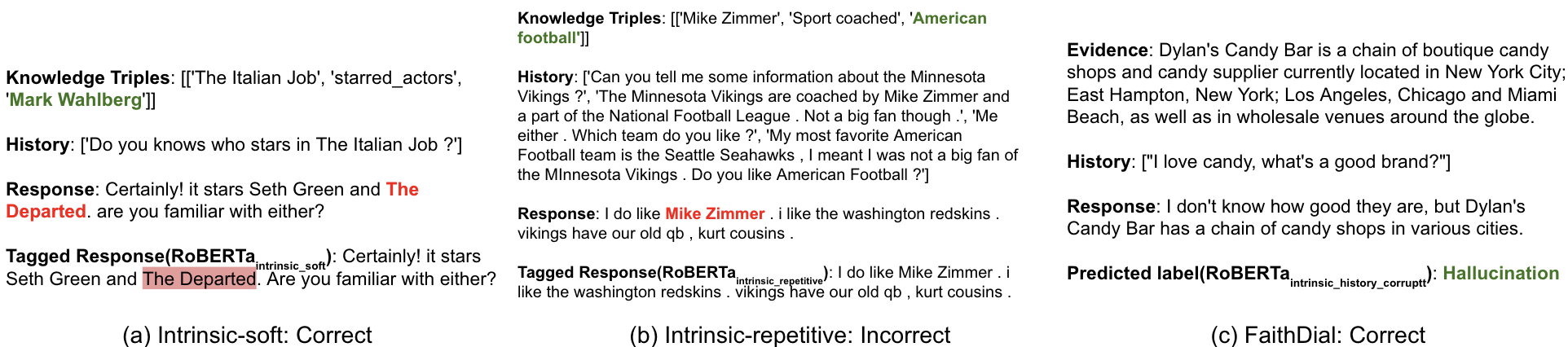}
    \caption{Positive and negative model predictions}
    \label{fig:example_viz}
\end{figure*}

\textbf{Baseline performance} Table \ref{tab:ind_results} and Table \ref{tab:main_results} show the baseline performance for the component datasets and mixed datasets. In both the settings, the utterance level hallucination classifier performs better than the token tagger in terms of F1. It can be inferred from Table  \ref{tab:ind_results} that, on average, it is comparatively easier to detect intrinsic hallucinations than extrinsic hallucinations; due to grounding on external knowledge, which indicates the validity of our perturbation techniques. However, comparing the occurrence statistics from Table \ref{tab:human_feedback}, it is noticed that extrinsic-soft hallucination, which has the least F1 score among all types, has the highest occurrences. In extrinsic-grouped and extrinsic-soft hallucinations, it is interesting that BERT performs better than the other pre-trained models. Now for mixed datasets, we ran inference on the test set of \textbf{observed} dataset, as expected F1 scores(for utterance classifier and token level tagger) of the  \textbf{observed} dataset are low as compared to other datasets due to high percentage of extrinsic-soft hallucination. Among other mixed datasets, the XLNet model fine-tuned on \textbf{extrinsic+} dataset performs best in terms of F1 scores. 
\begin{figure}[h]
\centering
\includegraphics[width=0.3\textwidth]{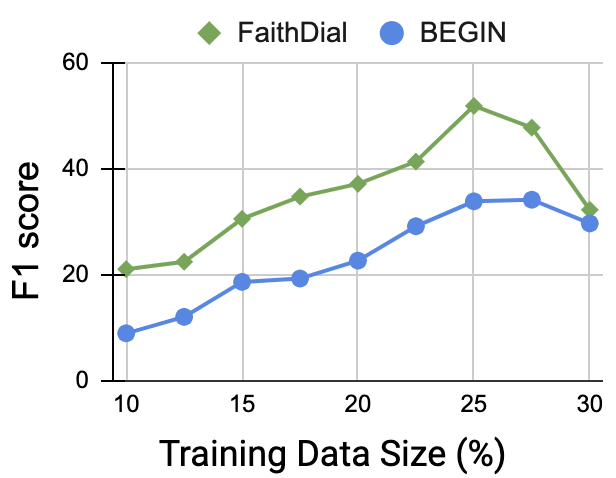}
\caption{Generalisation capability of RoBERTa-large model fine-tuned using multiple splits of intrinsic-history-corrupt dataset}
\label{fig:gen_graph}
\end{figure}

\textbf{Performance on human-verified data} We test the best performing models fine-tuned on our \textbf{mixed} datasets on human-verified data as described in \S \ref{sec:hu_veri}. Using the existing benchmark and baseline models, we also perform a zero-shot inference on the human-verified data. From Table \ref{tab:hu_data_performance}, it is clear that the models fine-tuned on existing benchmark data cannot understand fact hallucination, especially when entities are misplaced. On the other hand, models trained on our datasets have F1 scores over 90\% and outperform the current baseline by 10.16\% and 17.5\% in the two tasks using a pre-trained model with fewer parameters. This suggests that identifying abrupt fact hallucination is more challenging than other types of hallucination(like presenting more data than expected), which are more commonly exhibited in the benchmark datasets.
\begin{table}[]
\centering
\scalebox{0.4}{%
\begin{tabular}{@{}llll@{}}
\toprule
\textbf{Fine-tuned on} &
  \textbf{Pretrain Model} &
  \textbf{\begin{tabular}[c]{@{}l@{}}F1\\ (Utterance-level)\end{tabular}} &
  \textbf{\begin{tabular}[c]{@{}l@{}}F1\\ (Token-level)\end{tabular}} \\ \midrule
MNLI       & RoBERTa-large    & 12.5            & \multicolumn{1}{c}{-} \\
BEGIN      & RoBERTa-large    & 15.4            & \multicolumn{1}{c}{-} \\
FaithDial  & RoBERTa-large    & 22.1            & \multicolumn{1}{c}{-} \\ \midrule
Intrin-Extrin\cite{dziri-etal-2021-neural} & RoBERTa-large & 83.81 &  68.2        \\ \midrule
balanced   & RoBERTa-base     & 92.27           & 78.61                 \\
observed   & XLNet(base-cased)& 90.15           & 70.27                 \\
extrinsic+ & XLNet(base-cased)& \textbf{93.97*} & \textbf{85.7*}        \\
intrinsic+ & RoBERTa-base     & 93.01           & 84.33                 \\ \bottomrule
\end{tabular}}
\caption{Performance of several benchmark models and models trained on \textsc{FADE} on the 500 human-verified data( *p-value < 0.001))}
\label{tab:hu_data_performance}
\end{table}
\begin{table}[]
\centering
\scalebox{0.42}{%
\begin{tabular}{@{}llll@{}}
\toprule
\textbf{Fine-tuned on}         & \textbf{Model} & \textbf{BEGIN} & \textbf{FaithDial}    \\ \midrule
MNLI(3-way)\cite{begin}     & T5             & 49.5           & \multicolumn{1}{c}{-} \\
MNLI\cite{faithdial}        & RoBERTa-large  & 61.1           & 81.6                  \\ \midrule
intrinsic\_hard             & RoBERTa-base   & 37.12          & 51.34                 \\
intrinsic\_history\_corrupt & RoBERTa-base   & 43.23          & 63.11                 \\
intrinsic\_hard             & RoBERTa-large  & 44.42          & 64.1                  \\
intrinsic\_history\_corrupt & RoBERTa-large  & \ul{55.11}          & \ul{71.43}                 \\ \bottomrule
\end{tabular}}
\caption{Zero-sort inference F1 scores on BEGIN and FaithDial benchmarks using utterance classification models trained on \textsc{FADE}}
\label{tab:gen_perform}
\end{table}

\textbf{Generalisability} We make zero-shot inference on BEGIN and FaithDial datasets' test splits. To make a fair comparison with the benchmark models, we further fine-tune \texttt{roberta-large} model on our datasets. Table \ref{tab:gen_perform} shows that F1 scores obtained from our best models underperform the best performing baseline by 6\% in BEGIN dataset and 10.17\% in the FaithDial dataset. Even though the performance is low, we have to understand that the benchmark datasets contain hallucinations that are fundamentally very different from fact hallucinations. Also, we notice that models trained on intrinsic hallucination perform the best because the hallucinatory responses in the benchmark dataset do not deviate much from the evidence. To estimate how much training data is optimum for generalisability, we ran inference on benchmark datasets using models fine-tuned to 10\% to 30\% (with a step of 2.5\%) data in train split. As shown in Figure \ref{fig:gen_graph} approximately 25\% is found to be optimum. 

\textbf{Model Predictions}  We visualized the predictions on different datasets in Figure \ref{fig:example_viz}. Our models were able to easily identify the hallucinated entities as shown in Figure \ref{fig:example_viz}a here "The Departed" is a movie in which "Mark Wahlberg" has acted but is not related to the movie discussed in the context, i.e., "The Italian Job". Similarly, predictions made on the FaithDial dataset(Figure \ref{fig:example_viz}c) show that our models could produce accurate predictions when the response is generating something that is not expected, but the hallucination has similarities with the evidence. Our model sometimes fails to understand when the history is convoluted(Figure \ref{fig:example_viz}b)).

\section{Related Work}
\textbf{Hallucination in Dialogue Systems} Hallucination in knowledge-grounded dialogue generation system is an emerging area of research \cite{hall1, hall2, hall3, hall4, dziri-etal-2021-neural}. Prior work addressed this issue by conditioning generation on control tokens \cite{hall4}, by training a token level hallucination critic to identify troublesome entities and rectify them \cite{dziri-etal-2021-neural} or by augmenting a generative model with a knowledge retrieval mechanism \cite{hall3}. Though beneficial, these models are trained on noisy training data \cite{dziri2} which can amplify the hallucinations further. Closest to our work \cite{dziri-etal-2021-neural} has created a hallucination critic using extrinsic-intrinsic corruption strategies. In contrast, we create more fine-grained corruption strategies so that hallucinated data mimics the attributions of a neural chat module.

\textbf{Hallucination Evaluation} 
Recently several benchmarks have been introduced, such as BEGIN\cite{begin}, DialFact\cite{gupta-etal-2022-dialfact}, FaithDial\cite{faithdial} and Attributable to Identified Sources (AIS) \cite{ais} framework. Though these methods can serve as a decent benchmarking system, their performance in detecting entity-level hallucination is unknown. In this work, we further contribute to this problem by proposing an entity-level hallucination detector trained on data created by various fine-grained perturbation strategies.

\section{Conclusion}
In this work, we have analyzed the modes of entity-level fact hallucination, which is an open problem in KG-grounded dialogue systems. Through a human feedback analysis, we demonstrate that these KG-grounded neural generators manifest more nuanced hallucinations than straightforward studied approaches. We have proposed fine-grained perturbation strategies to create a dataset that mimics the real-world observations and create a series of datasets collectively known as \textsc{FADE}. Our entity-level hallucination detection model can predict hallucinated entities with an F1 score of 75.59\% and classify whether an utterance is hallucinated or not with an F1 score of 90.75\%. Our models can generalize well when zero-shot predictions are made on benchmarks like BEGIN and FaithDial, indicating our perturbation strategies' robustness. This work can be extended by devising more sophisticated perturbation mechanisms, which can simulate other types of hallucinations.

\section*{Limitations}

The major limitations of this work are as follows:
\begin{itemize}[noitemsep,nolistsep]
  \item The token-level hallucination classifier and utterance-level hallucination classifier can have contradictory results; however, this happens in a small percentage of data.
  \item Models trained on extrinsic datasets do not generalize well on the benchmark datasets, as the benchmark dataset contains hallucination mostly related to the evidence provided.
\end{itemize}

\section*{Acknowledgements}
We thank the anonymous reviewers for providing valuable feedback on our manuscript. This work is partly supported by NSF grant number IIS-2214070.  The content in this paper is solely the responsibility of the authors and does not necessarily represent the official views of the funding entity.

\bibliography{anthology,custom}

\begin{thebibliography}{24}
\expandafter\ifx\csname natexlab\endcsname\relax\def\natexlab#1{#1}\fi

\bibitem[{Bast et~al.(2014)Bast, B\"{a}urle, Buchhold, and
  Hau\ss{}mann}]{freebase}
Hannah Bast, Florian B\"{a}urle, Bj\"{o}rn Buchhold, and Elmar Hau\ss{}mann.
  2014.
\newblock \href {https://doi.org/10.1145/2567948.2577016} {Easy access to the
  freebase dataset}.
\newblock In \emph{Proceedings of the 23rd International Conference on World
  Wide Web}, WWW '14 Companion, page 95–98, New York, NY, USA. Association
  for Computing Machinery.

\bibitem[{Brown et~al.(2020)Brown, Mann, Ryder, Subbiah, Kaplan, Dhariwal,
  Neelakantan, Shyam, Sastry, Askell, Agarwal, Herbert-Voss, Krueger, Henighan,
  Child, Ramesh, Ziegler, Wu, Winter, Hesse, Chen, Sigler, Litwin, Gray, Chess,
  Clark, Berner, McCandlish, Radford, Sutskever, and Amodei}]{gpt3}
Tom~B. Brown, Benjamin Mann, Nick Ryder, Melanie Subbiah, Jared Kaplan,
  Prafulla Dhariwal, Arvind Neelakantan, Pranav Shyam, Girish Sastry, Amanda
  Askell, Sandhini Agarwal, Ariel Herbert-Voss, Gretchen Krueger, Tom Henighan,
  Rewon Child, Aditya Ramesh, Daniel~M. Ziegler, Jeffrey Wu, Clemens Winter,
  Christopher Hesse, Mark Chen, Eric Sigler, Mateusz Litwin, Scott Gray,
  Benjamin Chess, Jack Clark, Christopher Berner, Sam McCandlish, Alec Radford,
  Ilya Sutskever, and Dario Amodei. 2020.
\newblock \href {https://doi.org/10.48550/ARXIV.2005.14165} {Language models
  are few-shot learners}.

\bibitem[{Center(2005)}]{center2005brier}
NOAA-CIRES Climate~Diagnostics Center. 2005.
\newblock Brier skill scores, rocs, and economic value diagrams can report
  false skill.

\bibitem[{Dziri et~al.(2022{\natexlab{a}})Dziri, Kamalloo, Milton, Zaiane, Yu,
  Ponti, and Reddy}]{faithdial}
Nouha Dziri, Ehsan Kamalloo, Sivan Milton, Osmar Zaiane, Mo~Yu, Edoardo~M.
  Ponti, and Siva Reddy. 2022{\natexlab{a}}.
\newblock \href {https://doi.org/10.48550/ARXIV.2204.10757} {Faithdial: A
  faithful benchmark for information-seeking dialogue}.

\bibitem[{Dziri et~al.(2021{\natexlab{a}})Dziri, Madotto, Za{\"\i}ane, and
  Bose}]{dziri-etal-2021-neural}
Nouha Dziri, Andrea Madotto, Osmar Za{\"\i}ane, and Avishek~Joey Bose.
  2021{\natexlab{a}}.
\newblock \href {https://doi.org/10.18653/v1/2021.emnlp-main.168} {Neural path
  hunter: Reducing hallucination in dialogue systems via path grounding}.
\newblock In \emph{Proceedings of the 2021 Conference on Empirical Methods in
  Natural Language Processing}, pages 2197--2214, Online and Punta Cana,
  Dominican Republic. Association for Computational Linguistics.

\bibitem[{Dziri et~al.(2022{\natexlab{b}})Dziri, Milton, Yu, Zaiane, and
  Reddy}]{dziri2}
Nouha Dziri, Sivan Milton, Mo~Yu, Osmar Zaiane, and Siva Reddy.
  2022{\natexlab{b}}.
\newblock \href {https://doi.org/10.48550/ARXIV.2204.07931} {On the origin of
  hallucinations in conversational models: Is it the datasets or the models?}

\bibitem[{Dziri et~al.(2022{\natexlab{c}})Dziri, Milton, Yu, Zaiane, and
  Reddy}]{audit}
Nouha Dziri, Sivan Milton, Mo~Yu, Osmar Zaiane, and Siva Reddy.
  2022{\natexlab{c}}.
\newblock \href {https://doi.org/10.48550/ARXIV.2204.07931} {On the origin of
  hallucinations in conversational models: Is it the datasets or the models?}

\bibitem[{Dziri et~al.(2021{\natexlab{b}})Dziri, Rashkin, Linzen, and
  Reitter}]{begin}
Nouha Dziri, Hannah Rashkin, Tal Linzen, and David Reitter. 2021{\natexlab{b}}.
\newblock \href {https://doi.org/10.48550/ARXIV.2105.00071} {Evaluating
  groundedness in dialogue systems: The begin benchmark}.

\bibitem[{Esp{\'\i}ndola and Ebecken(2005)}]{espindola2005extending}
Rog{\'e}rio~P Esp{\'\i}ndola and Nelson~FF Ebecken. 2005.
\newblock On extending f-measure and g-mean metrics to multi-class problems.
\newblock \emph{WIT Transactions on Information and Communication
  Technologies}, 35.

\bibitem[{Gupta et~al.(2022)Gupta, Wu, Liu, and
  Xiong}]{gupta-etal-2022-dialfact}
Prakhar Gupta, Chien-Sheng Wu, Wenhao Liu, and Caiming Xiong. 2022.
\newblock \href {https://doi.org/10.18653/v1/2022.acl-long.263} {{D}ial{F}act:
  A benchmark for fact-checking in dialogue}.
\newblock In \emph{Proceedings of the 60th Annual Meeting of the Association
  for Computational Linguistics (Volume 1: Long Papers)}, pages 3785--3801,
  Dublin, Ireland. Association for Computational Linguistics.

\bibitem[{Maynez et~al.(2020)Maynez, Narayan, Bohnet, and
  McDonald}]{maynez-etal-2020-faithfulness}
Joshua Maynez, Shashi Narayan, Bernd Bohnet, and Ryan McDonald. 2020.
\newblock \href {https://doi.org/10.18653/v1/2020.acl-main.173} {On
  faithfulness and factuality in abstractive summarization}.
\newblock In \emph{Proceedings of the 58th Annual Meeting of the Association
  for Computational Linguistics}, pages 1906--1919, Online. Association for
  Computational Linguistics.

\bibitem[{Mielke et~al.(2020)Mielke, Szlam, Boureau, and Dinan}]{hall2}
Sabrina~J. Mielke, Arthur Szlam, Y-Lan Boureau, and Emily Dinan. 2020.
\newblock \href {https://doi.org/10.48550/ARXIV.2012.14983} {Linguistic
  calibration through metacognition: aligning dialogue agent responses with
  expected correctness}.

\bibitem[{Moon et~al.(2019)Moon, Shah, Kumar, and
  Subba}]{moon-etal-2019-opendialkg}
Seungwhan Moon, Pararth Shah, Anuj Kumar, and Rajen Subba. 2019.
\newblock \href {https://doi.org/10.18653/v1/P19-1081} {{O}pen{D}ial{KG}:
  Explainable conversational reasoning with attention-based walks over
  knowledge graphs}.
\newblock In \emph{Proceedings of the 57th Annual Meeting of the Association
  for Computational Linguistics}, pages 845--854, Florence, Italy. Association
  for Computational Linguistics.

\bibitem[{Parikh et~al.(2020)Parikh, Wang, Gehrmann, Faruqui, Dhingra, Yang,
  and Das}]{parikh-etal-2020-totto}
Ankur Parikh, Xuezhi Wang, Sebastian Gehrmann, Manaal Faruqui, Bhuwan Dhingra,
  Diyi Yang, and Dipanjan Das. 2020.
\newblock \href {https://doi.org/10.18653/v1/2020.emnlp-main.89} {{ToTTo}: A
  controlled table-to-text generation dataset}.
\newblock In \emph{Proceedings of the 2020 Conference on Empirical Methods in
  Natural Language Processing (EMNLP)}, pages 1173--1186, Online. Association
  for Computational Linguistics.

\bibitem[{Radford et~al.(2019)Radford, Wu, Child, Luan, Amodei, Sutskever
  et~al.}]{gpt2}
Alec Radford, Jeffrey Wu, Rewon Child, David Luan, Dario Amodei, Ilya
  Sutskever, et~al. 2019.
\newblock Language models are unsupervised multitask learners.
\newblock \emph{OpenAI blog}, 1(8):9.

\bibitem[{Rashkin et~al.(2021{\natexlab{a}})Rashkin, Nikolaev, Lamm, Collins,
  Das, Petrov, Tomar, Turc, and Reitter}]{ais}
Hannah Rashkin, Vitaly Nikolaev, Matthew Lamm, Michael Collins, Dipanjan Das,
  Slav Petrov, Gaurav~Singh Tomar, Iulia Turc, and David Reitter.
  2021{\natexlab{a}}.
\newblock \href {https://doi.org/10.48550/ARXIV.2112.12870} {Measuring
  attribution in natural language generation models}.

\bibitem[{Rashkin et~al.(2021{\natexlab{b}})Rashkin, Reitter, Tomar, and
  Das}]{hall4}
Hannah Rashkin, David Reitter, Gaurav~Singh Tomar, and Dipanjan Das.
  2021{\natexlab{b}}.
\newblock \href {https://doi.org/10.18653/v1/2021.acl-long.58} {Increasing
  faithfulness in knowledge-grounded dialogue with controllable features}.
\newblock In \emph{Proceedings of the 59th Annual Meeting of the Association
  for Computational Linguistics and the 11th International Joint Conference on
  Natural Language Processing (Volume 1: Long Papers)}, pages 704--718, Online.
  Association for Computational Linguistics.

\bibitem[{Roller et~al.(2021)Roller, Dinan, Goyal, Ju, Williamson, Liu, Xu,
  Ott, Smith, Boureau, and Weston}]{hall1}
Stephen Roller, Emily Dinan, Naman Goyal, Da~Ju, Mary Williamson, Yinhan Liu,
  Jing Xu, Myle Ott, Eric~Michael Smith, Y-Lan Boureau, and Jason Weston. 2021.
\newblock \href {https://doi.org/10.18653/v1/2021.eacl-main.24} {Recipes for
  building an open-domain chatbot}.
\newblock In \emph{Proceedings of the 16th Conference of the European Chapter
  of the Association for Computational Linguistics: Main Volume}, pages
  300--325, Online. Association for Computational Linguistics.

\bibitem[{See and Manning(2021)}]{see-manning-2021-understanding}
Abigail See and Christopher Manning. 2021.
\newblock \href {https://aclanthology.org/2021.sigdial-1.1} {Understanding and
  predicting user dissatisfaction in a neural generative chatbot}.
\newblock In \emph{Proceedings of the 22nd Annual Meeting of the Special
  Interest Group on Discourse and Dialogue}, pages 1--12, Singapore and Online.
  Association for Computational Linguistics.

\bibitem[{Shuster et~al.(2021)Shuster, Poff, Chen, Kiela, and Weston}]{hall3}
Kurt Shuster, Spencer Poff, Moya Chen, Douwe Kiela, and Jason Weston. 2021.
\newblock \href {https://doi.org/10.18653/v1/2021.findings-emnlp.320}
  {Retrieval augmentation reduces hallucination in conversation}.
\newblock In \emph{Findings of the Association for Computational Linguistics:
  EMNLP 2021}, pages 3784--3803, Punta Cana, Dominican Republic. Association
  for Computational Linguistics.

\bibitem[{Smith et~al.(2022)Smith, Hsu, Qian, Roller, Boureau, and
  Weston}]{smith-etal-2022-human}
Eric Smith, Orion Hsu, Rebecca Qian, Stephen Roller, Y-Lan Boureau, and Jason
  Weston. 2022.
\newblock \href {https://doi.org/10.18653/v1/2022.nlp4convai-1.8} {Human
  evaluation of conversations is an open problem: comparing the sensitivity of
  various methods for evaluating dialogue agents}.
\newblock In \emph{Proceedings of the 4th Workshop on NLP for Conversational
  AI}, pages 77--97, Dublin, Ireland. Association for Computational
  Linguistics.

\bibitem[{Tuan et~al.(2019)Tuan, Chen, and Lee}]{tuan-etal-2019-dykgchat}
Yi-Lin Tuan, Yun-Nung Chen, and Hung-yi Lee. 2019.
\newblock \href {https://doi.org/10.18653/v1/D19-1194} {{D}y{K}g{C}hat:
  Benchmarking dialogue generation grounding on dynamic knowledge graphs}.
\newblock In \emph{Proceedings of the 2019 Conference on Empirical Methods in
  Natural Language Processing and the 9th International Joint Conference on
  Natural Language Processing (EMNLP-IJCNLP)}, pages 1855--1865, Hong Kong,
  China. Association for Computational Linguistics.

\bibitem[{Wiseman et~al.(2017)Wiseman, Shieber, and
  Rush}]{wiseman-etal-2017-challenges}
Sam Wiseman, Stuart Shieber, and Alexander Rush. 2017.
\newblock \href {https://doi.org/10.18653/v1/D17-1239} {Challenges in
  data-to-document generation}.
\newblock In \emph{Proceedings of the 2017 Conference on Empirical Methods in
  Natural Language Processing}, pages 2253--2263, Copenhagen, Denmark.
  Association for Computational Linguistics.

\bibitem[{Wolf et~al.(2020)Wolf, Debut, Sanh, Chaumond, Delangue, Moi, Cistac,
  Rault, Louf, Funtowicz, Davison, Shleifer, von Platen, Ma, Jernite, Plu, Xu,
  Le~Scao, Gugger, Drame, Lhoest, and Rush}]{wolf-etal-2020-transformers}
Thomas Wolf, Lysandre Debut, Victor Sanh, Julien Chaumond, Clement Delangue,
  Anthony Moi, Pierric Cistac, Tim Rault, Remi Louf, Morgan Funtowicz, Joe
  Davison, Sam Shleifer, Patrick von Platen, Clara Ma, Yacine Jernite, Julien
  Plu, Canwen Xu, Teven Le~Scao, Sylvain Gugger, Mariama Drame, Quentin Lhoest,
  and Alexander Rush. 2020.
\newblock \href {https://doi.org/10.18653/v1/2020.emnlp-demos.6} {Transformers:
  State-of-the-art natural language processing}.
\newblock In \emph{Proceedings of the 2020 Conference on Empirical Methods in
  Natural Language Processing: System Demonstrations}, pages 38--45, Online.
  Association for Computational Linguistics.

\end{thebibliography}
\bibliographystyle{acl_natbib}

\appendix

\section{Definition Details}
\label{sec:def_details}
\begin{enumerate}[(a), wide, labelwidth=!, labelindent=0pt]
  \begin{figure}[h]
    \centering
    \includegraphics[width=0.3\textwidth]{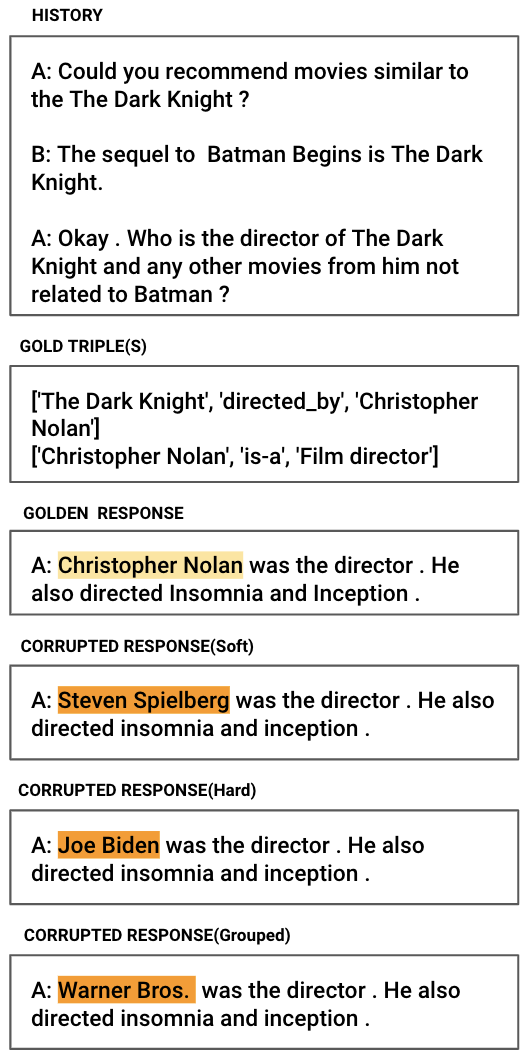}
    \caption{Extrinsic Hallucination}%
    \label{ext_hal}
  \end{figure}
  
  \item (Extrinsic-Soft). \textit{An extrinsic-soft hallucination corresponds to an utterance that brings a new span of text which is similar to the expected span but does not correspond to a valid triple in  $ \mathcal{G}^{k}_{c}$.}
  
  A hallucination is considered extrinsic when knowledge is injected which is not authentically captured by $ \mathcal{G}^{k}_{c}$. However, the injected knowledge is similar to the expected entity. Identifying this type of hallucination can be challenging due to the high similarity between the injected and gold knowledge. For example, in Figure \ref{ext_hal} the dialogue sample contains an extrinsic-soft hallucination as the entity in response -- "Steven Spielberg" is similar to "Christopher Nolan", and it is not supported within 1-hop sub-graph.
  
  \item (Extrinsic-Hard). \textit{An extrinsic-hard hallucination corresponds to an utterance that brings a new span of text which is different from the expected span and does not correspond to a valid triple in  $ \mathcal{G}^{k}_{c}$.}
  
  An extrinsic-hard hallucination occurs when injected knowledge is dissimilar to the expected entity and is not supported within the 1-hop sub-graph. It is easier to detect extrinsic-hard than extrinsic-soft as the entities are fundamentally different from the entities present in the 1-hop sub-graph. However, the entity type is retained, like an entity with a type "person" will be replaced by the same type of entity. Figure \ref{ext_hal} shows an example of extrinsic-hard hallucination, where the golden entity "Christopher Nolan" is replaced by a different category of entity, "Joe Biden", but the type of entity is retained.
  
  \item (Extrinsic-Grouped). \textit{An extrinsic-grouped hallucination corresponds to an utterance that brings a new span of text which is different from the expected span but is of a specific predefined type and does not correspond to a valid triple in  $ \mathcal{G}^{k}_{c}$.}
  
  Like an extrinsic-hard hallucination, extrinsic-grouped hallucination introduces an entity that is functionally different from the original entity and not supported by the 1-hop sub-graph. The only difference is that the corrupted entity is not of the same type; instead, it is replaced by an entity of a similar type, defined in Table \ref{tab:ent_group}. For example, Figure \ref{ext_hal} shows "Christopher Nolan" which is of type "person" is replaced by "Warner Bros." of type "organization". Here, the types "person" and "organization" are placed in the same group.
  
  \begin{table}[]
    \centering
    \scalebox{0.45}{%
    \begin{tabular}{@{}lll@{}}
    \toprule
    Group & Definition                                          & Groups                            \\ \midrule
    1 &
      \begin{tabular}[c]{@{}l@{}}A person, organization, political party, or part \\ of a religious group can be related to each other.\end{tabular} &
      "PERSON", "ORG", "NORP \\
    2 &
      \begin{tabular}[c]{@{}l@{}}Location, building, airports,   infrastructure\\ elements, countries, cities, and states can be interrelated\end{tabular} &
      "LOC", "GPE", "FAC" \\
    3     & A product, work of art, or law can be interrelated. & "PRODUCT", "WORK\_OF\_ART", "LAW" \\ \bottomrule
    \end{tabular}}
    \caption{Defined groups for extrinsic-grouped hallucination}
    \label{tab:ent_group}
  \end{table}
  
  \begin{figure}[h]
    \centering
    \includegraphics[width=0.3\textwidth]{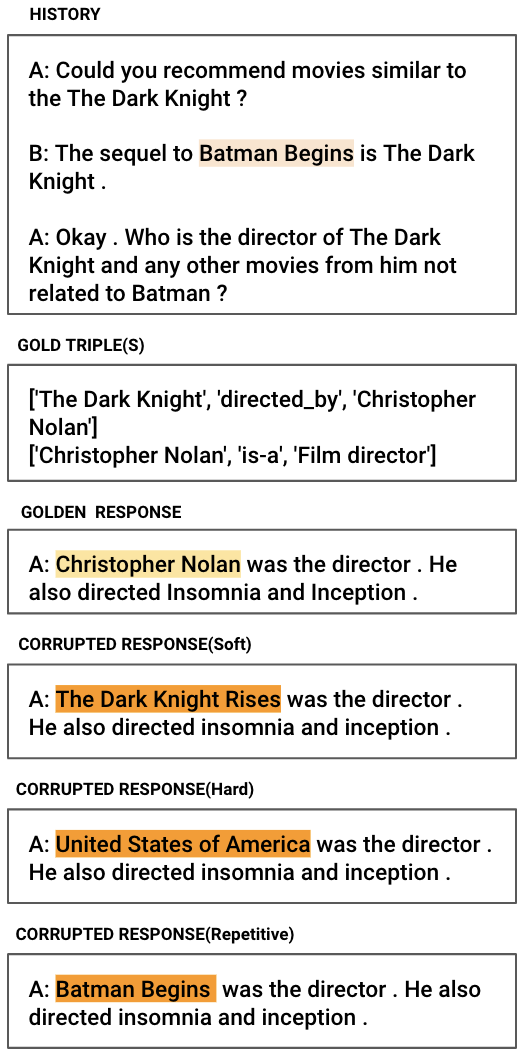}
    \caption{Intrinsic Hallucination}%
    \label{int_hal}
  \end{figure}
  
  \item (Intrinsic-Soft). \textit{An intrinsic-soft hallucination corresponds to an utterance that misuses any triple in $ \mathcal{G}^{k}_{c}$ such that there is no direct path between the entities, but they are similar to each other.}
  
  Intrinsic hallucinations occur when the KG triples are misused, especially in intrinsic-soft hallucination an entity is selected from $ \mathcal{G}^{k}_{c}$ which is very similar or closely related to the original entity. For example, in Figure \ref{int_hal}, "Christopher Nolan" is replaced with "The Dark Knight Rises" which is retrieved from the 1-hop sub-graph and has close relation with the original entity "Christopher Nolan".

  \item (Intrinsic-Hard). \textit{An intrinsic-hard hallucination corresponds to an utterance that misuses any triple in $ \mathcal{G}^{k}_{c}$ such that there is no direct path between the entities, and they are not related in any form.}
  
  Similar to intrinsic-soft hallucination, it also misuses the information in KG triples. However, the similarity of the corrupted entity with the original entity is relatively tiny. For example, in Figure \ref{int_hal}, "Christopher Nolan" is replaced with "United States of America". Although the corrupted entity is drawn from  $\mathcal{G}^{k}_{c}$, it is very different from the original entity.
  
  \item (Intrinsic-Repetitive). \textit{An intrinsic-repetitive hallucination corresponds to an utterance that either misuse \texttt{[SBJ]} or \texttt{[OBJ]} in $ \mathcal{G}^{k}_{c}$ such that there is no direct path between the entities but the entity has previously occurred in conversational history.}. 
  
  An entity from the conversational history is often repeated in the current utterances, which corresponds to intrinsic-repetitive hallucination. Here, an entity from the history which also occurs in $\mathcal{G}^{k}_{c}$ and of high relatedness, is swapped with the original entity. Figure \ref{int_hal} shows "Batman Begins" which is supported by $\mathcal{G}^{k}_{c}$ is replaced with "Christopher Nolan".
   
  \begin{figure}[h]
    \centering
    \includegraphics[width=0.3\textwidth]{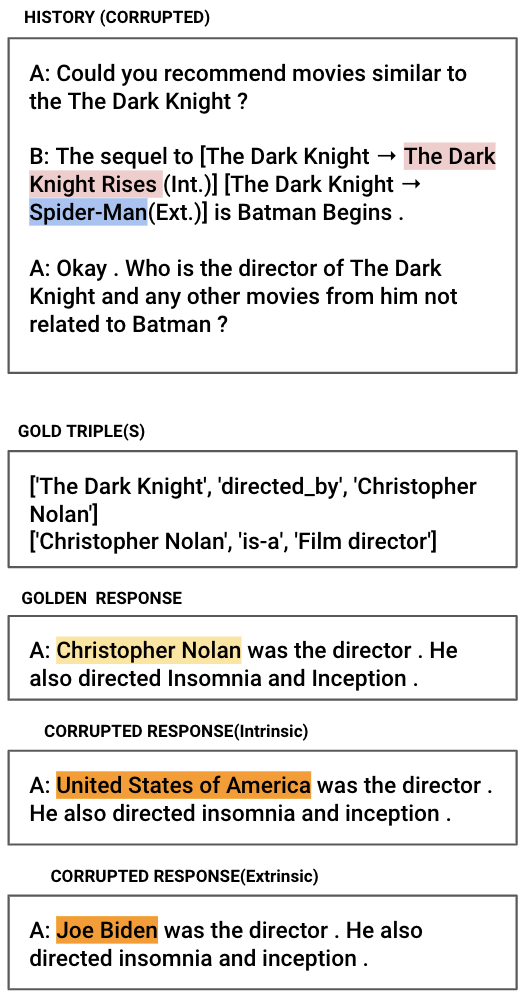}
    \caption{History Corrupted Hallucination}%
    \label{his_cor_hal}
  \end{figure}
  
  \item (History Corrupted- Intrinsic/ Extrinsic). \textit{A history corrupted(intrinsic/extrinsic) hallucination corresponds to an utterance subjected to intrinsic or extrinsic hallucination influenced by hallucinated entities in conversational history.}
  
  Sometimes conversational agents are driven into a perplexed state, and we can witness hallucinations in most turns. So, this hallucinated history can trigger hallucination in the current utterance. This phenomenon can be seen both in extrinsic and intrinsic forms of hallucination. Figure \ref{his_cor_hal} depicts extrinsic/intrinsic hallucination occurring in history -- "The Dark Knight" is changed to "The Dark Knight Rises" for intrinsic hallucination; similarly, "The Dark Knight" is changed to "Spider-Man" for extrinsic hallucination. Hallucinations in the current utterance happen as described in previous sections.
  
\end{enumerate}

\section{AMT Instructions} \label{sec:amt_ins}
\begin{figure*}
  \centering
    \includegraphics[width=0.7\textwidth]{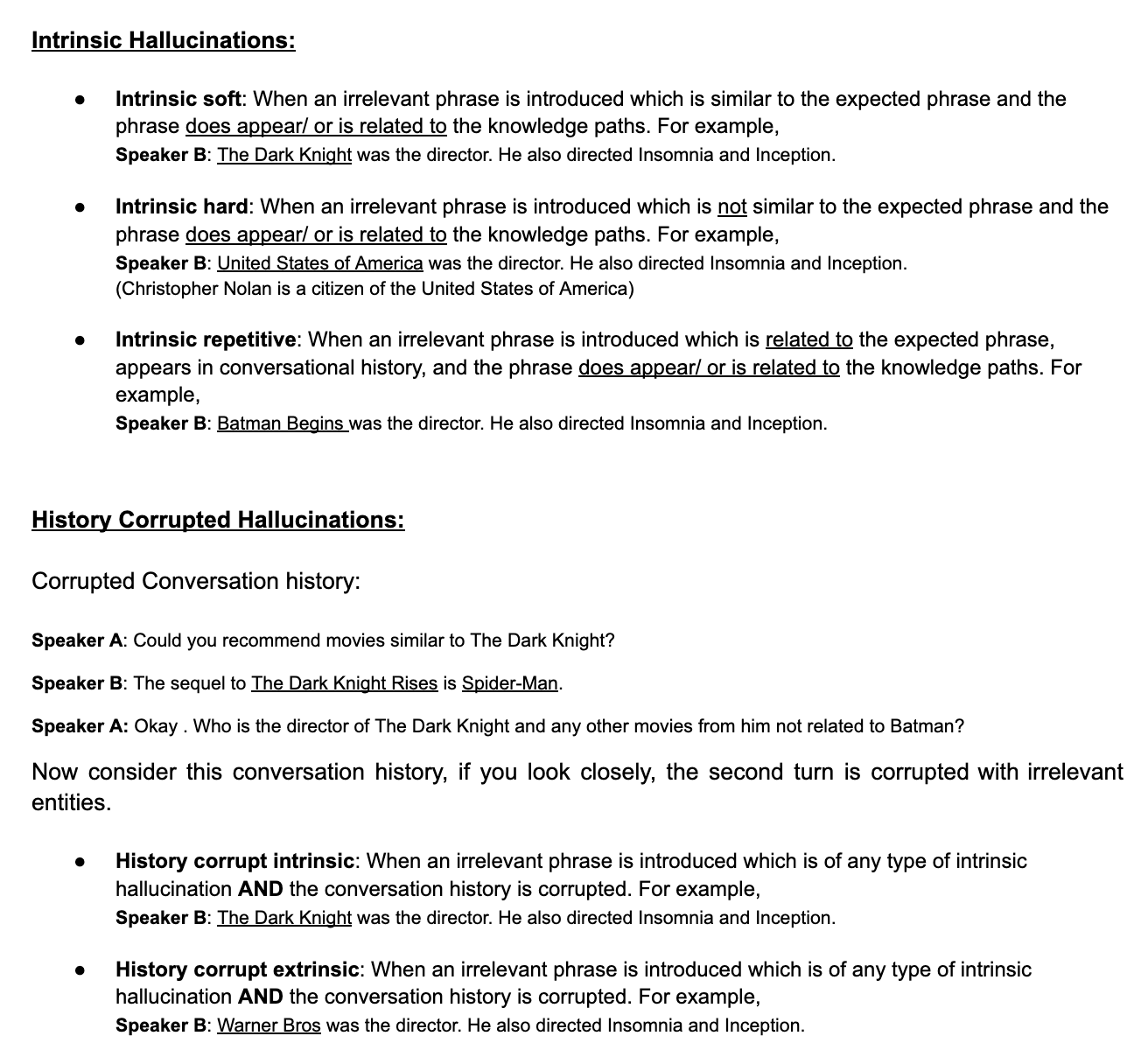}
    \caption{Annotation interface for human feedback analysis(Instructions, part 2)}
    \label{fig:amt2}
\end{figure*}

\begin{figure*}
  \centering
    \includegraphics[width=0.7\textwidth]{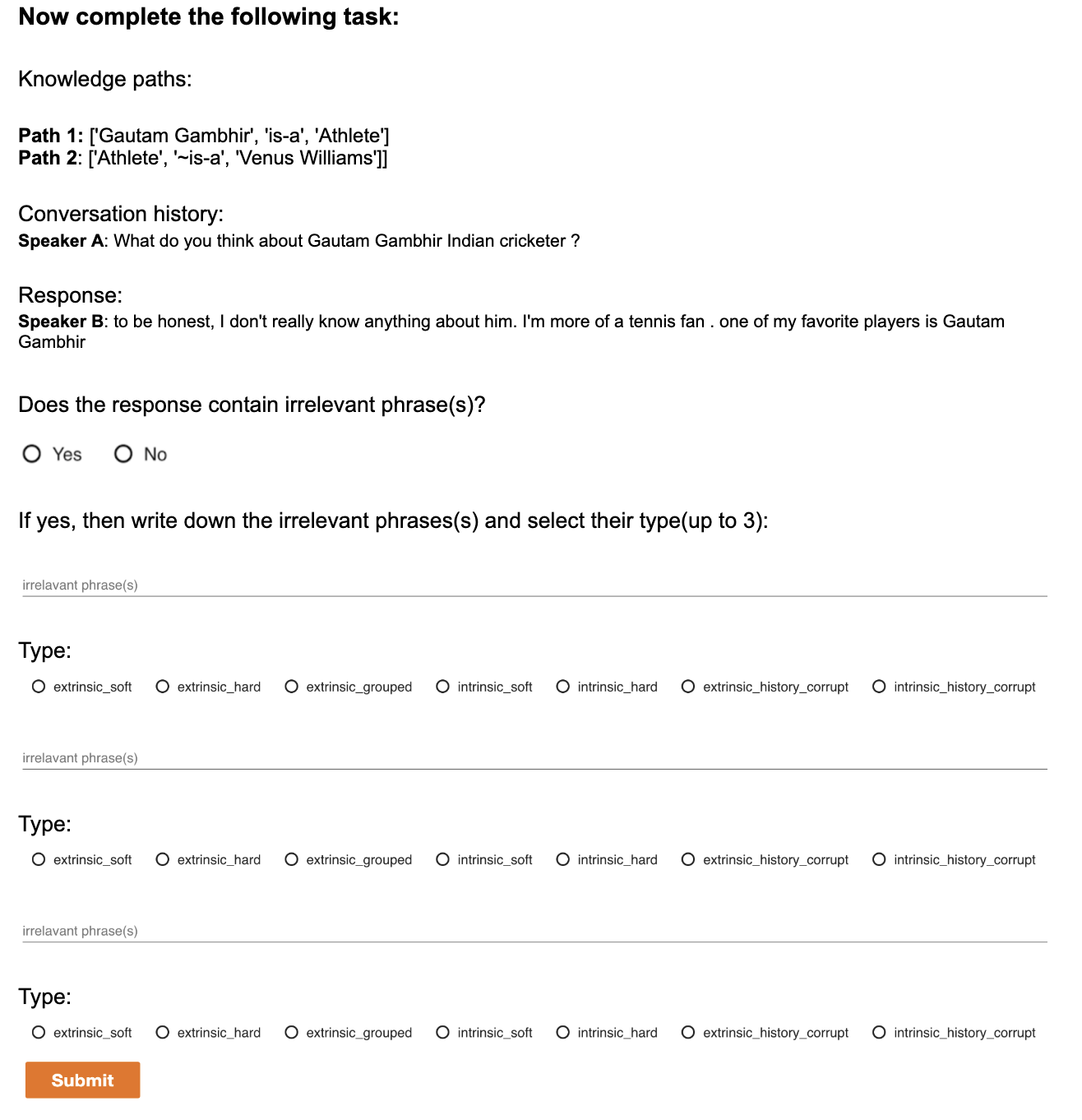}
    \caption{Annotation interface for human feedback analysis(example annotation, workers were ask to find up to 3 spans if hallucinations are found in the data)}
    \label{fig:amt3}
\end{figure*}

We present the screenshot of the annotation interface in Figure \ref{fig:amt2}, \ref{fig:amt2} and  \ref{fig:amt3}. Workers were paid an average of \$7-8 per hour across all tasks. We agree that this annotation process has a high learning curve. Even workers with high approval rates made errors in the initial rounds of annotation. A graduate computer science student manually verified randomly selected samples and provided feedback to the workers. Feedback was given to the workers, especially when they selected the same answers for ten consecutive HITS. After sending feedback three times, all spammed HITS were discarded. 

\begin{figure*}
  \centering
    \includegraphics[width=0.7\textwidth]{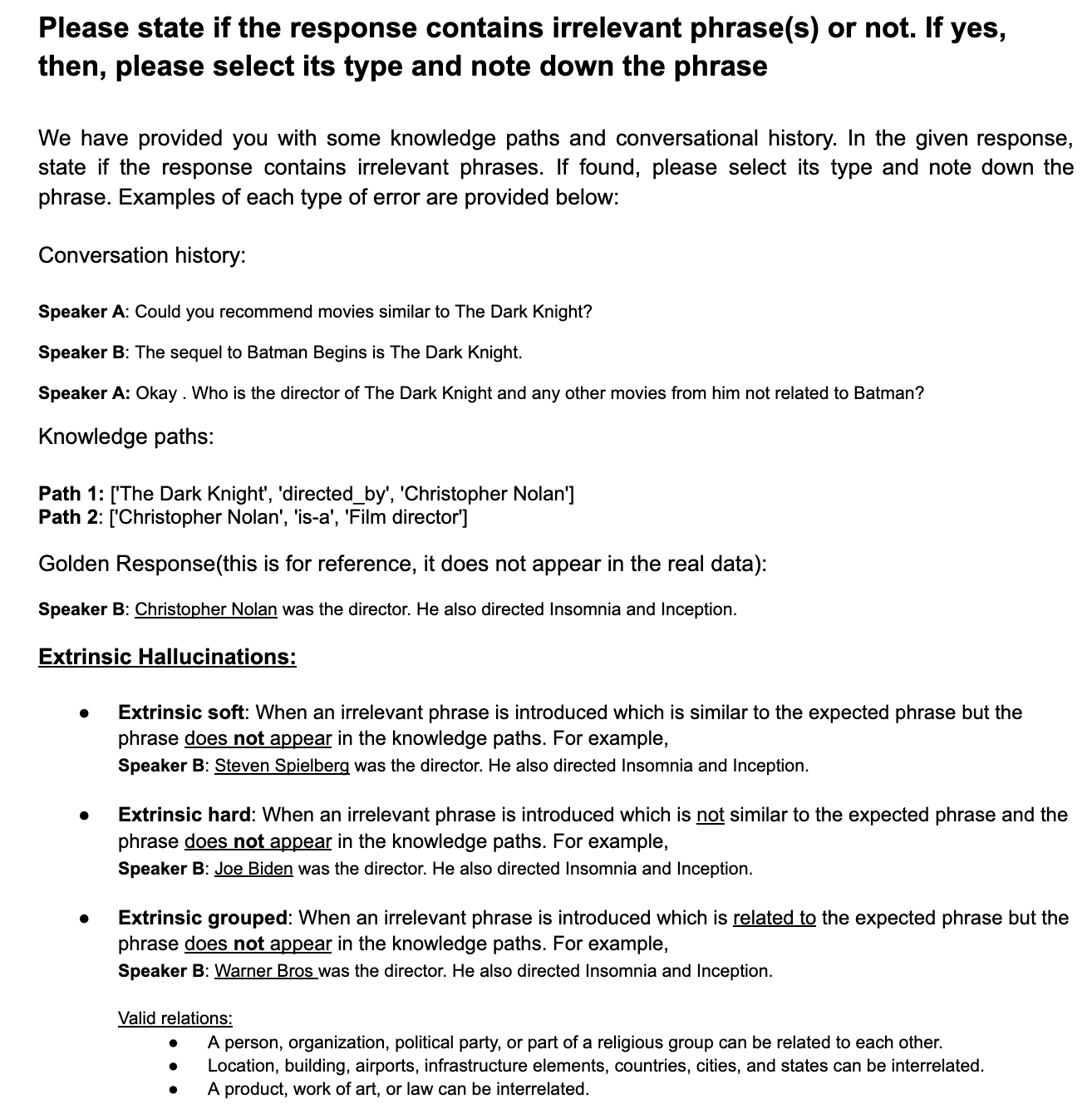}
    \caption{Annotation interface for human feedback analysis(Instructions, part 1)}
    \label{fig:amt1}
\end{figure*}

\section{OpenDialKG} \label{opendialkg}

We use OpenDialKG \cite{moon-etal-2019-opendialkg}, a crowded-sourced English dialogue dataset where two workers are paired together to chat about a particular topic. The first speaker is requested to start the conversation about a given entity. The second speaker is assigned to write an accurate response based on facts extracted from an existing KG, Freebase \cite{freebase}. The facts represent paths from the KG that are either 1-hop or 2-hop from the initial entity. Once the second speaker responds, the first speaker continues discussing the topic engagingly, and new multi-hop facts from the KG are shown to the second speaker. The dialogue can be considered as traversing multiple paths in the KG. However, not all utterances within the same conversation are grounded on facts from the KG. The second speaker can decide not to select a path from the KG to form an answer and instead forms a "chit-chat" response. Overall, the dataset consists of four domains: movie, music, sport, and book, where each second speaker's utterance is annotated with paths from the KG. The KG corresponds to an extensive subgraph extracted from Freebase with $\sim$ 1.2M triples (subject, predicate, object), $\sim$ 101k distinct entities, and 1357 distinct relations. We use 77,430 data points in the dataset for constructing \textsc{FADE}. 

\section{Perturbation Hyper-parameters}
\subsection{Search Index Details}
We use \texttt{Solr} in case of extrinsic hallucination. We use the BM25 index, defined by the class \texttt{solr.BM25SimilarityFactory}. We manually labeled 50 data points(for the entity type PERSON) for tuning the indexes through grid search. Grid-search conditions were as follows: b was varied from 0.3 to 0.9 with a step of 0.1 and k1 was varied from 0.8 to 2.0 with a step of 0.2. Following grid search, an optimum MAP score of 0.789 was found, with b = 0.9 and k1= 1.6. For the dynamic indexes that were created in the case of intrinsic hallucination, we use the python library \url{https://github.com/dorianbrown/rank_bm25} with default configurations.
\subsection{Free parameter \& $\beta$ optimization} \label{int_perturb}
We use a free term weight parameter($\varepsilon $) in intrinsic hallucination to represent the queries and nodes. Similar to extrinsic hallucination we manually annotated 50 data-points and ran grid search for $\varepsilon \in \{10^{-i}, 2\times10^{-i}; i\in\{1,5\}\}$, and found $\varepsilon=2\times10^{-4}$ to be the optimum value. We used the same technique for optimizing $\beta$, and the search space ranged from 0.1 to 0.7 with a step of 0.05. 
\subsection{KG embeddings}\label{kg_emb}
We follow the same approach \cite{dziri-etal-2021-neural} for generating the KG embeddings. OpenDialKG triples are also represented using a textual term called "render". For the triples containing this term, we pass it through to GPT2 and then extract hidden state representations for each entity’s word piece and finally obtain a final representation by applying a MaxPool over the hidden representations. For entity mentions not described in “render”, we get their representations directly from the last hidden states in GPT2.

\subsection{Mixing Ratios}\label{mix_ratio}
Mixing ratios for creating the mixed datasets are defined in Table \ref{tab:mixdata}. Perturbed and non-perturbed samples are drawn randomly from component datasets.
\begin{table}[]
\centering
\scalebox{0.3}{%
\begin{tabular}{@{}llllllllll@{}}
\toprule
\textbf{Dataset   Type} &
  \textbf{Ext-Soft(\%)} &
  \textbf{Ext-Hard(\%)} &
  \textbf{Ext-Grp(\%)} &
  \textbf{Int-Soft(\%)} &
  \textbf{Int-Hard(\%)} &
  \textbf{Int-Rep(\%)} &
  \textbf{HC-Ext(\%)} &
  \textbf{HC-Int(\%)} &
  \textbf{N-Halluc(\%)} \\ \midrule
Observed   & 12.495 & 6.4425 & 1.04  & 0.92  & 1.025 & 1.7   & 2.4575 & 1.4575 & 72.4625 \\
Balanced   & 6.25   & 6.25   & 6.25  & 6.25  & 6.25  & 6.25  & 6.25   & 6.25   & 50      \\
Extrinsic+ & 12.5   & 9.375  & 9.375 & 6.25  & 6.25  & 6.25  & 6.25   & 6.25   & 37.5    \\
Intrinsic+ & 6.25   & 6.25   & 6.25  & 9.375 & 9.375 & 9.375 & 6.25   & 6.25   & 40.625  \\ \bottomrule
\end{tabular}}
\caption{Mixing ratios for different datasets}
\label{tab:mixdata}
\end{table}

\section{Implementation Details} \label{impl_det}

The utterance and token level classifier are implemented using the Pytorch Huggingface Transformers library \cite{wolf-etal-2020-transformers}. The following configuration were found to be best performing for each models, as shown in Table \ref{tab:rbase}, \ref{tab:rlarge}, \ref{tab:bert} and \ref{tab:xlnet}. The models were trained in a single NVIDIA A5000 GPU, the average running time for the base models were 2.5 hours, and for the large model was $\sim$ 5 hours.

\begin{table}[]
\centering
\scalebox{0.6}{%
\begin{tabular}{@{}ll@{}}
\toprule
\textbf{Hyperparameter}       & \textbf{Value} \\ \midrule
train\_batch\_size            & 12             \\
gradient\_accumulation\_steps & 2              \\
num\_train\_epochs            & 4(Token)/10(Utt)             \\
weight\_decay                 & 0.01           \\
warmup\_proportion            & 0.1            \\
learning\_rate                & 1e-5           \\
adam\_epsilon                 & 1e-8           \\
max\_grad\_norm               & 1              \\
eval\_batch\_size             & 18             \\ \bottomrule
\end{tabular}}
\caption{RoBERTa-base hyper parameters}
\label{tab:rbase}
\end{table}

\begin{table}[]
\centering
\scalebox{0.6}{%
\begin{tabular}{@{}ll@{}}
\toprule
\textbf{Hyperparameter}       & \textbf{Value} \\ \midrule
train\_batch\_size            & 12             \\
gradient\_accumulation\_steps & 2              \\
num\_train\_epochs            & 4(Token)/10(Utt)               \\
weight\_decay                 & 0.01           \\
warmup\_proportion            & 0.1            \\
learning\_rate                & 2e-5           \\
adam\_epsilon                 & 1.5e-8           \\
max\_grad\_norm               & 1              \\
eval\_batch\_size             & 18             \\ \bottomrule
\end{tabular}}
\caption{RoBERTa-large hyper parameters}
\label{tab:rlarge}
\end{table}

\begin{table}[]
\centering
\scalebox{0.6}{%
\begin{tabular}{@{}ll@{}}
\toprule
\textbf{Hyperparameter}       & \textbf{Value} \\ \midrule
train\_batch\_size            & 12             \\
gradient\_accumulation\_steps & 2              \\
num\_train\_epochs            & 4(Token)/10(Utt)               \\
weight\_decay                 & 0.01           \\
warmup\_proportion            & 0.1            \\
learning\_rate                & 5e-5           \\
adam\_epsilon                 & 1e-8           \\
max\_grad\_norm               & 1              \\
eval\_batch\_size             & 18             \\ \bottomrule
\end{tabular}}
\caption{BERT-base-uncased hyper parameters}
\label{tab:bert}
\end{table}

\begin{table}[]
\centering
\scalebox{0.6}{%
\begin{tabular}{@{}ll@{}}
\toprule
\textbf{Hyperparameter}       & \textbf{Value} \\ \midrule
train\_batch\_size            & 12             \\
gradient\_accumulation\_steps & 2              \\
num\_train\_epochs            & 4(Token)/10(Utt)        \\
weight\_decay                 & 0.01           \\
warmup\_proportion            & 0.1            \\
learning\_rate                & 5e-5           \\
adam\_epsilon                 & 1e-8           \\
max\_grad\_norm               & 1              \\
eval\_batch\_size             & 18             \\ \bottomrule
\end{tabular}}
\caption{XLNet-base hyper parameters}
\label{tab:xlnet}
\end{table}

\section{Supplementary results}

We report metrics for all the models trained using 25\% of the dataset, for component datasets in Table \ref{tab:all_ind_results} and mixed datasets in Table \ref{tab:all_main_results}.
\begin{table*}[t]
\centering
\resizebox{14cm}{!}{%
\begin{tabular}{@{}l|l|lll|llllll@{}}
\toprule
\multicolumn{1}{c|}{\multirow{2}{*}{\textbf{Dataset}}} &
  \multicolumn{1}{c|}{\multirow{2}{*}{\textbf{Best Model}}} &
  \multicolumn{3}{c|}{\textbf{Token Level}} &
  \multicolumn{6}{c}{\textbf{Utterance Level}} \\ \cmidrule(l){3-11} 
\multicolumn{1}{c|}{} &
  \multicolumn{1}{c|}{} &
  \multicolumn{1}{c}{\textbf{F1}} &
  \multicolumn{1}{c}{\textbf{P}} &
  \multicolumn{1}{c|}{\textbf{R}} &
  \multicolumn{1}{c}{\textbf{F1}} &
  \multicolumn{1}{c}{\textbf{P}} &
  \multicolumn{1}{c}{\textbf{R}} &
  \multicolumn{1}{c}{\textbf{G-Mean}} &
  \multicolumn{1}{c}{\textbf{BSS}} &
  \multicolumn{1}{c}{\textbf{AUC}} \\ \midrule
extrinsic\_hard             & roberta-base      & 0.70613382 & 0.68956357 & 0.72352004 & 0.86181139 & 0.83985441 & 0.88494727 & 0.93029609 & 0.03277357 & 0.93145803 \\
extrinsic\_grouped          & roberta-base      & 0.7986706  & 0.77534593 & 0.82344214 & 0.90499405 & 0.89090483 & 0.91953606 & 0.93487266 & 0.0589842  & 0.93500056 \\
intrinsic\_hard             & roberta-base      & 0.84409519 & 0.84717262 & 0.84104004 & 0.90789771 & 0.92741563 & 0.8891844  & 0.93192336 & 0.04522725 & 0.9329505  \\
intrinsic\_soft             & roberta-base      & 0.78797921 & 0.80193163 & 0.774504   & 0.87102229 & 0.90540109 & 0.83915877 & 0.90255779 & 0.06217348 & 0.90495271 \\
intrinsic\_repetitive       & roberta-base      & 0.82702178 & 0.82759578 & 0.82644857 & 0.88005638 & 0.89506881 & 0.86553923 & 0.92305012 & 0.03146957 & 0.92496078 \\
intrinsic\_history\_corrupt & roberta-base      & 0.83406626 & 0.82763636 & 0.84059684 & 0.90857229 & 0.92340555 & 0.89420804 & 0.93381877 & 0.04511612 & 0.93469609 \\
extrinsic\_history\_corrupt & roberta-base      & 0.72010547 & 0.71212516 & 0.72826667 & 0.87400219 & 0.85486834 & 0.89401217 & 0.93612638 & 0.02971357 & 0.93711831 \\
extrinsic\_soft             & roberta-base      & 0.60045426 & 0.60811376 & 0.59298532 & 0.72017689 & 0.74873563 & 0.69371672 & 0.81231271 & 0.09344656 & 0.82245014 \\
extrinsic\_hard             & bert-base-uncased & 0.71146832 & 0.72259569 & 0.70067846 & 0.88285121 & 0.88299233 & 0.88271013 & 0.93232489 & 0.02705296 & 0.93371925 \\
extrinsic\_grouped          & bert-base-uncased & 0.80688364 & 0.8056026  & 0.80816875 & 0.91302235 & 0.9180408  & 0.90805848 & 0.93577473 & 0.05285693 & 0.93619772 \\
intrinsic\_hard             & bert-base-uncased & 0.83328471 & 0.82308025 & 0.84374538 & 0.91416629 & 0.92395896 & 0.90457903 & 0.93917074 & 0.04259417 & 0.93983215 \\
intrinsic\_soft             & bert-base-uncased & 0.75277325 & 0.79087205 & 0.71817644 & 0.85483616 & 0.91836735 & 0.79952621 & 0.88349437 & 0.06794858 & 0.88790364 \\
intrinsic\_repetitive       & bert-base-uncased & 0.7481198  & 0.71392596 & 0.78575388 & 0.84134941 & 0.82295256 & 0.86058758 & 0.91436157 & 0.04330141 & 0.9160416  \\
intrinsic\_history\_corrupt & bert-base-uncased & 0.82318199 & 0.8229997  & 0.82336435 & 0.90891209 & 0.9316067  & 0.8872969  & 0.93164021 & 0.04459424 & 0.93274826 \\
extrinsic\_history\_corrupt & bert-base-uncased & 0.67029785 & 0.69294369 & 0.64908533 & 0.87358552 & 0.88214169 & 0.86519372 & 0.92312672 & 0.02886461 & 0.9250663  \\
extrinsic\_soft             & bert-base-uncased & 0.64089366 & 0.6922167  & 0.59665579 & 0.7480315  & 0.81958894 & 0.68796592 & 0.81616138 & 0.08033967 & 0.82810534 \\
extrinsic\_hard             & xlnet-base-cased  & 0.72115512 & 0.71982018 & 0.72249502 & 0.8736255  & 0.8712651  & 0.87599872 & 0.92800607 & 0.02926739 & 0.92954989 \\
extrinsic\_grouped          & xlnet-base-cased  & 0.78452923 & 0.77288925 & 0.79652518 & 0.89920345 & 0.8915677  & 0.90697112 & 0.92895654 & 0.06212166 & 0.92922301 \\
intrinsic\_hard             & xlnet-base-cased  & 0.84443122 & 0.85082459 & 0.83813322 & 0.90878914 & 0.92875867 & 0.88966027 & 0.93238499 & 0.04477944 & 0.93341088 \\
intrinsic\_soft             & xlnet-base-cased  & 0.76722735 & 0.80484632 & 0.73296801 & 0.85379657 & 0.90941058 & 0.80459259 & 0.88491991 & 0.06892207 & 0.88892969 \\
intrinsic\_repetitive       & xlnet-base-cased  & 0.7941989  & 0.79135701 & 0.79706127 & 0.86978508 & 0.88154897 & 0.85833102 & 0.91820183 & 0.03428001 & 0.9202899  \\
intrinsic\_history\_corrupt & xlnet-base-cased  & 0.83667247 & 0.82269807 & 0.85112982 & 0.91298209 & 0.91864812 & 0.90738552 & 0.9396723  & 0.04337198 & 0.94024672 \\
extrinsic\_history\_corrupt & xlnet-base-cased  & 0.72378159 & 0.72354039 & 0.72402294 & 0.88100942 & 0.87862377 & 0.88340807 & 0.93239789 & 0.02749991 & 0.93375627 \\
extrinsic\_soft             & xlnet-base-cased  & 0.60896216 & 0.63207547 & 0.58747961 & 0.73844753 & 0.79296016 & 0.69094782 & 0.81535862 & 0.08484401 & 0.82655921 \\ \bottomrule
\end{tabular}%
}
\caption{All models benchmark (numbers in fractions) for component datasets, models trained on 25\% of the total dataset.}
\label{tab:all_ind_results}
\end{table*}

\begin{table*}[t]
\centering
\resizebox{12cm}{!}{%
\begin{tabular}{@{}l|l|lll|llllll@{}}
\toprule
\multicolumn{1}{c|}{\multirow{2}{*}{\textbf{Dataset}}} &
  \multicolumn{1}{c|}{\multirow{2}{*}{\textbf{Best Model}}} &
  \multicolumn{3}{c|}{\textbf{Token Level}} &
  \multicolumn{6}{c}{\textbf{Utterance Level}} \\ \cmidrule(l){3-11} 
\multicolumn{1}{c|}{} &
  \multicolumn{1}{c|}{} &
  \multicolumn{1}{c}{\textbf{F1}} &
  \multicolumn{1}{c}{\textbf{P}} &
  \multicolumn{1}{c|}{\textbf{R}} &
  \multicolumn{1}{c}{\textbf{F1}} &
  \multicolumn{1}{c}{\textbf{P}} &
  \multicolumn{1}{c}{\textbf{R}} &
  \multicolumn{1}{c}{\textbf{G-Mean}} &
  \multicolumn{1}{c}{\textbf{BSS}} &
  \multicolumn{1}{c}{\textbf{AUC}} \\ \midrule
balanced        & roberta-base      & 0.73405875 & 0.68751809 & 0.78735795 & 0.882424   & 0.83853553 & 0.93116042 & 0.86213807 & 0.131385   & 0.86469621 \\
observed        & roberta-base      & 0.62554537 & 0.59004757 & 0.66558773 & 0.77904114 & 0.73266454 & 0.83168565 & 0.85077041 & 0.14126728 & 0.85098938 \\
extrinsic\_plus & roberta-base      & 0.74849152 & 0.71339648 & 0.78721816 & 0.90921175 & 0.87804878 & 0.94266814 & 0.84332203 & 0.12278872 & 0.84855698 \\
intrinsic\_plus & roberta-base      & 0.75045075 & 0.71112613 & 0.79437919 & 0.90157054 & 0.86518353 & 0.94115257 & 0.84511316 & 0.12778319 & 0.85001331 \\
balanced        & bert-base-uncased & 0.6570643  & 0.57930535 & 0.7589345  & 0.85119497 & 0.78285516 & 0.9326075  & 0.81309735 & 0.17265032 & 0.82075474 \\
observed        & bert-base-uncased & 0.59965325 & 0.52847854 & 0.6929832  & 0.76124302 & 0.67629046 & 0.87060443 & 0.84589508 & 0.16352531 & 0.84624573 \\
extrinsic\_plus & bert-base-uncased & 0.72993044 & 0.663004   & 0.81188563 & 0.90179749 & 0.84940317 & 0.96108049 & 0.8086632  & 0.1365238  & 0.82074909 \\
intrinsic\_plus & bert-base-uncased & 0.71653573 & 0.65640721 & 0.78879093 & 0.89301716 & 0.84373548 & 0.94841293 & 0.82126564 & 0.14130552 & 0.82978853 \\
balanced        & xlnet-base-cased  & 0.71863497 & 0.66214437 & 0.78566356 & 0.87222741 & 0.81850039 & 0.93350331 & 0.84619893 & 0.14481173 & 0.85028143 \\
observed        & xlnet-base-cased  & 0.63436089 & 0.57976023 & 0.70031519 & 0.77706573 & 0.71053723 & 0.85733951 & 0.8540217  & 0.14730018 & 0.85402812 \\
extrinsic\_plus & xlnet-base-cased  & 0.75593757 & 0.7079124  & 0.81095307 & 0.90747949 & 0.86768256 & 0.95110254 & 0.83209459 & 0.12649216 & 0.8395401  \\
intrinsic\_plus & xlnet-base-cased  & 0.74488988 & 0.68995602 & 0.80932808 & 0.90141776 & 0.85748704 & 0.95009285 & 0.83869733 & 0.129222   & 0.84522772 \\ \bottomrule
\end{tabular}%
}
\caption{All model benchmark (numbers in fractiom) for mixed datasets, models trained on 25\% of the total dataset.}
\label{tab:all_main_results}
\end{table*}




\end{document}